\documentclass[10pt,twocolumn,letterpaper]{article}

\usepackage{iccv}
\usepackage{times}
\usepackage{epsfig}
\usepackage{graphicx}
\usepackage{amsmath}
\usepackage{amssymb}
\usepackage{booktabs}
\usepackage{multirow}
\usepackage{caption}
\usepackage{subcaption}
\usepackage[table,xcdraw]{xcolor}

\usepackage{amsmath,amsfonts,bm}

\def\eqref#1{equation~\ref{#1}}

\def\1{\bm{1}}

\def\vg{{\bm{g}}}

\def\vw{{\bm{w}}}

\def\mM{{\bm{M}}}

\def\mW{{\bm{W}}}

\DeclareMathAlphabet{\mathsfit}{\encodingdefault}{\sfdefault}{m}{sl}
\SetMathAlphabet{\mathsfit}{bold}{\encodingdefault}{\sfdefault}{bx}{n}

\def\gD{{\mathcal{D}}}

\def\gS{{\mathcal{S}}}
\def\gT{{\mathcal{T}}}

\def\sR{{\mathbb{R}}}

\usepackage{algorithm}
\usepackage{algorithmic}
\usepackage{rotating}

\newlength\savewidth
  
  \makeatletter
\def\@fnsymbol#1{\ensuremath{\ifcase#1\or \dagger\or \ddagger\or
   \mathsection\or \mathparagraph\or \|\or **\or \dagger\dagger
   \or \ddagger\ddagger \else\@ctrerr\fi}}
\makeatother

\definecolor{citecolor}{HTML}{0071bc}
\usepackage[pagebackref=true,breaklinks=true,letterpaper=true,colorlinks,  citecolor=citecolor,bookmarks=false]{hyperref}

\usepackage{xcolor}
\newcommand{\bohan}[1]{\textcolor{black}{#1}}

\iccvfinalcopy 

\def\iccvPaperID{2240} 

\ificcvfinal\fi

\usepackage{colortbl} 
\usepackage{graphicx} 
\usepackage{enumitem}  
\usepackage{xspace}  

\newcommand{\SPT}[0]{\textsc{SPT}}

\newcommand{\SPTa}[0]{\textsc{SPT-Adapter}}
\newcommand{\SPTl}[0]{\textsc{SPT-LoRA}}
\newcommand{\vprompt}[0]{\textsc{Prompt}}
\newcommand{\deepprompt}[0]{\vprompt{}-\textsc{deep}}
\newcommand{\shallowprompt}[0]{\vprompt{}-\textsc{shallow}}

\newcommand{\partialft}[0]{\textsc{Partial}}
\newcommand{\linear}[0]{\textsc{Linear}}
\newcommand{\fullft}[0]{\textsc{Full}}

\newcommand{\mlp}[0]{\textsc{Mlp}}
\newcommand{\bias}[0]{\textsc{Bias}}
\newcommand{\adapter}[0]{\textsc{Adapter}}
\newcommand{\adaptformer}[0]{\textsc{AdaptFormer}}
\newcommand{\lora}[0]{\textsc{LoRA}}
\newcommand{\noah}[0]{\textsc{NOAH}}

\newcommand{\vit}[0]{ViT}

\newcommand{\moco}[0]{MoCo v3}

\newcommand{\mae}[0]{MAE}

\newcommand{\imagenet}[0]{ImageNet}

\newcommand{\cub}[0]{CUB-200-2011}
\newcommand{\nabirds}[0]{NABirds}
\newcommand{\flowers}[0]{Oxford Flowers}
\newcommand{\cars}[0]{Stanford Cars}
\newcommand{\dogs}[0]{Stanford Dogs}
\newcommand{\vtab}[0]{VTAB-1k}

\definecolor{tabvline}{HTML}{a8a495}
\definecolor{prompt_blue}{HTML}{1f78b4}
\definecolor{prompt_red}{HTML}{d45c43}

\definecolor{green_im}{rgb}{0.0, 0.5, 0.0}

\newcommand{\band}{\rowcolor{gray!15}}

\definecolor{mypink}{RGB}{255,105,180}
\def\rev{\textcolor{black}}

\begin{document}
\title{Sensitivity-Aware Visual Parameter-Efficient Fine-Tuning}

\author{%
  Haoyu He$^{1}$ ~~ Jianfei Cai$^{1}$ ~~ Jing Zhang$^{2}$  ~~ Dacheng Tao$^{2}$ ~~ Bohan Zhuang$^{1}\thanks{Corresponding author. E-mail: $\tt  bohan.zhuang@gmail.com$}$ \\ [0.25cm]
$^1$ ZIP Lab, Monash University \quad $^2$ The University of Sydney \\[0.1cm]
}

\maketitle
\ificcvfinal\thispagestyle{empty}\fi

\begin{abstract}
Visual Parameter-Efficient Fine-Tuning (PEFT) has become a powerful alternative for full fine-tuning so as to adapt pre-trained vision models to downstream tasks, which only tunes a small number of parameters while freezing the vast majority ones to ease storage burden and optimization difficulty. However, existing PEFT methods introduce trainable parameters to the same positions across different tasks depending solely on human heuristics and neglect the domain gaps. To this end, we study where to introduce and how to allocate trainable parameters by proposing a novel \textbf{S}ensitivity-aware visual \textbf{P}arameter-efficient fine-\textbf{T}uning (SPT) scheme, which adaptively allocates trainable parameters to task-specific important positions given a desired tunable parameter budget. Specifically, our SPT first quickly identifies the sensitive parameters that require tuning for a given task in a data-dependent way. Next, our SPT further boosts the representational capability for the weight matrices whose number of sensitive parameters exceeds a pre-defined threshold by utilizing existing structured tuning methods, e.g., LoRA~\cite{hu2022lora} or Adapter~\cite{houlsby2019parameter}, 
to replace directly tuning the selected sensitive parameters (unstructured tuning) under the budget. Extensive experiments on a wide range of downstream recognition tasks show that our SPT is complementary to the existing PEFT methods and largely boosts their performance, e.g., SPT improves Adapter with supervised pre-trained ViT-B/16 backbone by 4.2\% and 1.4\% mean Top-1 accuracy, reaching SOTA performance on FGVC and VTAB-1k benchmarks, respectively. Source code is at \url{https://github.com/ziplab/SPT}.
\end{abstract}

\section{Introduction}

\begin{figure}[t]
\begin{center}    \includegraphics[width=0.95\linewidth]{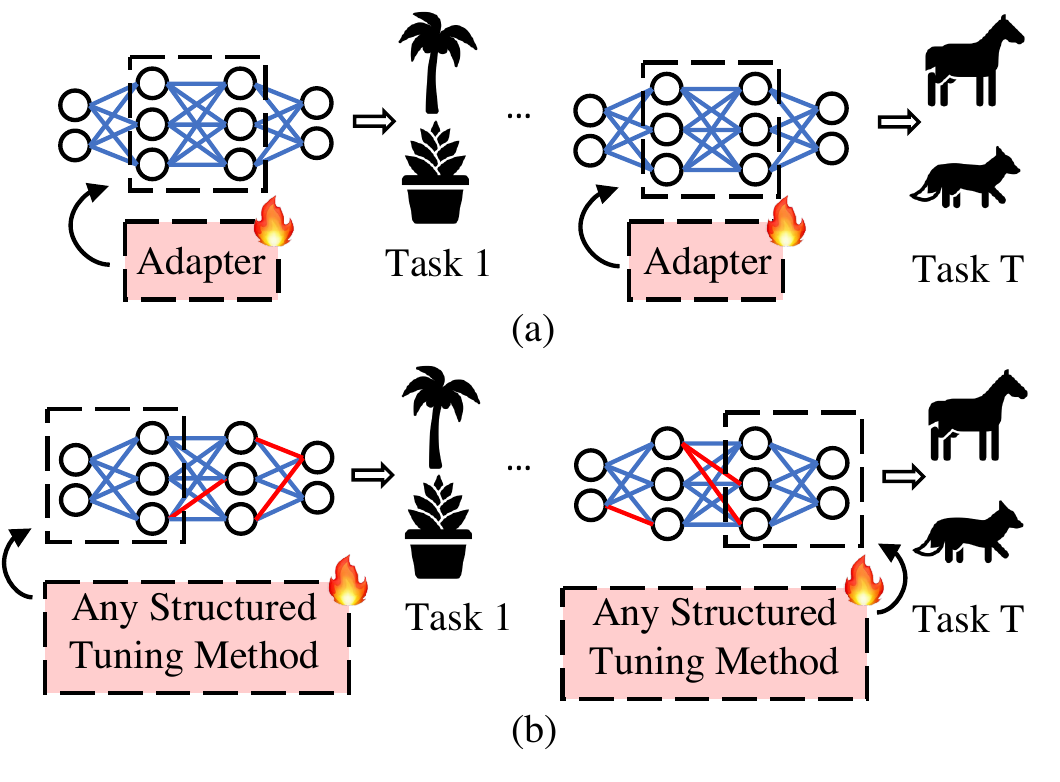}
\end{center}\vspace{-2em}
\caption{(a) Existing PEFT methods, such as Adapter~\cite{houlsby2019parameter} introduce trainable parameters to the same positions for all downstream tasks. These methods design task-agnostic positions to employ trainable parameters relying on heuristics and neglect consideration of the distinct domain gaps and characteristics for the downstream tasks. (b) Our Sensitivity-aware visual Parameter-efficient fine-Tuning (SPT) introduces trainable parameters to the task-specific important positions and allocates them with both unstructured and structured tuning granularities, simultaneously. For structured tuning, SPT can exploit any existing structured tuning methods, such as LoRA~\cite{hu2022lora} or Adapter~\cite{houlsby2019parameter}. Red lines and blocks represent trainable parameters and modules, while blue lines represent frozen parameters.}
\label{fig:intro}
\vspace{-1.5em}
\end{figure}

To effectively adapt the pre-trained representations to the downstream tasks, the de-facto choice is full fine-tuning, which initializes the model with the pre-trained weights and tunes all the parameters. 
However, vanilla full fine-tuning needs to store a separate instance of parameters for each task and each deployment scenario. It can be extremely storage-intensive as the storage cost grows linearly with the number of possible cases, considering there are vast varieties of downstream tasks and dynamic deployment environments, especially when deploying the large vision models~\cite{vit,swin,xu2021vitae} to mobile systems. For example, even storing a single large pre-trained ViT-H~\cite{vit} model on a local disk requires at least 2.3GB, while the Top-10 U.S. apps required only collectively 2.2GB in May 2021.\footnote{https://sensortower.com/blog/ios-app-size-growth-2021}

Notably, an emerging solution is to replace vanilla fine-tuning with visual Parameter-Efficient Fine-Tuning (PEFT)~\cite{jia2022vpt,chen2022adaptformer,zhang2022neural,jie2022convolutional}, which only tunes a small number of trainable parameters while freezing the vast majority ones that are shared by multiple tasks.
As PEFT approaches exhibit less than 1\% of trainable parameters, the storage burden is largely alleviated. Another attractive property of PEFT is that tuning fewer parameters eases the optimization difficulty and mitigates the overfitting issue when adapting large pre-trained models on the target dataset, thereby achieving comparable or even better performance than vanilla fine-tuning~\cite{jia2022vpt}. Although promising, the existing PEFT approaches introduce trainable parameters to the same positions for all downstream tasks, relying on human heuristics and neglecting task-specific domain gaps and characteristics, which limits their performance. For instance, in a task-agnostic manner, Prompt Tuning-deep~\cite{jia2022vpt} and Adapter~\cite{houlsby2019parameter} respectively add trainable parameters to multi-head self-attention and feed-forward network layers for all distinct tasks as depicted in Figure~\ref{fig:intro} (a).

To address this fundamental challenge,
we explore \textit{where to introduce} and \textit{how to allocate} trainable parameters 
under a desired parameter budget by presenting a novel \textbf{S}ensitivity-aware visual \textbf{P}arameter-efficient fine-\textbf{T}uning (SPT) scheme that identifies the \textit{task-specific important positions} to adaptively allocate trainable parameters. Since the pre-trained weights at distinct positions 
have varying contributions for 
different downstream tasks~\cite{yosinski2014transferable,kumar2022finetuning,neyshabur2020being}, we first propose a new criterion to quickly identify the task-specific sensitive parameters that require tuning in a data-dependent way. Inspired by model pruning metrics~\cite{srivastava2015training,molchanov2019importance,bender2018understanding,brock2017smash}, we propose to measure the parameter sensitivity with the loss reduction when being tuned, which can be
approximated by a first-order Taylor expansion derived within a single forward and backward pass ahead of fine-tuning in one-shot. Our sensitivity criterion is simple and effective, which can identify the task-specific important positions to introduce trainable parameters for any backbone quickly. For instance, calculating the sensitivity for ViT-B/16 backbone takes only 5.5 seconds with a single GPU on any of the \vtab{} datasets.

With our criterion, we empirically observe that the proportions of the sensitivity parameters for each block indeed vary markedly across different tasks in Section~\ref{subsec:observation}. To allocate the trainable parameters under a desired trainable parameter budge, an intuitive solution is to directly tune the most sensitive weight connections, which we name unstructured tuning. Despite its simplicity and flexibility, unstructured tuning only tunes a few parameters which still lacks representational capability and is challenging to bridge the domain gap. To this end, we propose to further incorporate structured tuning to replace unstructured tuning at the sensitive weight matrices whose numbers of sensitive parameters exceed a pre-defined threshold to improve the representational capability under a similar parameter budget. Structured tuning can be implemented by any parameter-efficient structured tuning methods~\cite{hu2022lora,chen2022adaptformer,jie2022convolutional,jia2022vpt} that directly adjust the hidden representations, \eg, inserting an adapter module sequentially after the sensitive weight matrices. Therefore, our SPT adaptively combines both unstructured and structured tuning granularity and allocates trainable parameters with high flexibility and representational capability for each distinct downstream task.

This paper has the following key contributions. 1) We make the pioneering exploration to identify the task-specific important positions under the PEFT setting, which is fast, effective, versatile to be applied to various backbones with different pre-training strategies, and orthogonal to the existing PEFT methods.
2) Based on the sensitivity criterion, we propose a trainable parameter allocation strategy that adaptively combines both unstructured and structured tuning under a desired parameter budget to achieve high flexibility, large capacity, and favorable trade-off between parameter efficiency and accuracy.
3) Extensive experiments on a total of 24 downstream recognition tasks with both plain and hierarchical vision Transformer backbones under supervised and self-supervised pre-trainings show that our SPT is complementary to the existing PEFT methods and boosts their performance by large margins. For instance, SPT improves Adapter~\cite{houlsby2019parameter} by 4.2\% mean Top-1 accuracy, outperforming the SOTA PEFT methods on the FGVC benchmark.
\vspace{-0.5em}
\section{Related Work}
\label{subsec:related}
\vspace{-0.3em}

\noindent{\textbf{Parameter-efficient fine-tuning.}} 
Full fine-tuning is the most predominant approach when adapting a large-scale pre-trained model to downstream tasks, where the model is initialized from the pre-trained weights with all parameters trainable. Yet, when a model becomes larger, parameter-efficient fine-tuning~\cite{lester2021power,li-liang-2021-prefix} is highly desirable, which tunes only a tiny portion of parameters to alleviate the storage burden. The general PEFT approaches can be categorized into addition-based PEFT methods and reparameterization-based PEFT methods. 

\emph{Addition-based PEFT} attaches additional trainable parameters to the backbone and only tunes these parameters. Apart from Prompt tuning~\cite{jia2022vpt} and Adapter~\cite{houlsby2019parameter}, recent addition-based methods study connecting or combining existing PEFT methods. For instance, He \etal~\cite{he2022towards} connect Prompt tuning and Adapter and provide a unified view that all PEFT approaches share the same design to adjust the hidden representations. Zhang \etal~\cite{zhang2022neural} search for the optimal configurations to combine multiple PEFT approaches following once-for-all scheme~\cite{cai2019once,wu2021autoformer}. Since the additional parameters require extra computations compared to full fine-tuning, a few recent works~\cite{sung2022lst,tu2022visual} design specific architectures to avoid storing the intermediate activations, thereby alleviating the fine-tuning memory cost. However, it is noteworthy that enhancing training efficiency is not the primary objective of our work.

\emph{Reparameterization-based PEFT} aims to avoid extra computational costs by tuning parameters that are inherently in or can be reparameterized into the backbone during inference. Prior works select the parameters that are inherently in the backbone, including the bias terms~\cite{zaken2022bitfit}, the last several layers~\cite{yosinski2014transferable,caelles2017one}, and weight connections~\cite{guo2021parameter,zhao2020masking}. To reparameterize new parameters into the backbone~\cite{hao2023consolidator,lian2022scaling}, representative work LoRA~\cite{hu2022lora} optimizes two low-rank matrices which can be further merged into the weight matrices. In contrast to the aforementioned works, we argue the importance of tuning parameters at task-specific important positions and quickly identify them with our proposed parameter sensitivity criterion before tuning, which is complementary to and provides valuable guidance for the existing PEFT methods. Moreover, our SPT can also be inference-efficient when implementing structured tuning with any reparameterization-based structured tuning method. Recently, SSF~\cite{lian2022scaling} is proposed to introduce trainable scaling and shifting parameters that can be absorbed into the previous linear layers. However, it cannot scale to higher trainable parameter budgets and requires a complex and time-consuming hyper-parameter search for learning rate, weight decay, and drop-path rate on each individual dataset, thus is not directly comparable to our method.

\noindent\textbf{Task-specific transfer learning.} The effectiveness of transferring pre-trained models to downstream tasks strongly depends on the relationship between the source and target tasks~\cite{rosenstein2005transfer,wang2019characterizing,kumar2022finetuning,plested2022deep}. This has motivated the community to explore the optimal pre-training data~\cite{cui2018large,yoon2020data}, model~\cite{tran2019transferability,nguyen2020leep}, and weights~\cite{guo2019spottune,xu2021raise} for the target task. To seek suitable \emph{task-specific pre-training data}, Cui~\etal~\cite{cui2018large} select the source domain data from the top-k most similar classes measured by Earth Mover's Distance; Yoon~\etal~\cite{yoon2020data} weight each class in the source domain with reinforcement learning; and Puigcerver~\etal~\cite{puigcerver2020scalable} first train a diverse set of experts and then select the most relevant expert for each target task. Another line of work selects a suitable \emph{pre-trained model for the target task} ahead of fine-tuning by measuring the transferability of pre-trained models to the target domain with interclass covariance between the source data and target classes~\cite{bao2019information} or conditional cross-entropy~\cite{tran2019transferability} between the source and target labels. Considering the transferability of the feature representations at distinct layers for the same pre-trained model is different~\cite{yosinski2014transferable,neyshabur2020being}, recent works~\cite{guo2020adafilter,sun2020adashare} endeavour \emph{transfer task-specific weights} by freezing some pre-trained weights and fine-tuning the rest.
For example, the task-specific fine-tuned weights are selected by learning a policy network with Gumbel-Softmax~\cite{guo2019spottune}, optimizing a sparse mask with $L_0$ norm~\cite{guo2021parameter}, and learning binary gates for each parameter~\cite{zhao2020masking}. Our SPT also adaptively selects task-specific parameters. In contrast to the previous work, we 1) derive task-specific important positions prior to fine-tuning with only a single forward and backward pass, which is computationally efficient; 2) mask the gradients for insensitive parameters in unstructured tuning with fixed binary masks, thereby having more affordable fine-tuning memory than optimizing learnable binary masks in~\cite{guo2021parameter,zhao2020masking}. Moreover, we are pioneering work to adaptively allocate task-specific trainable parameters with both fine-grained unstructured and coarse-grained structured tuning granularities to achieve both high flexibility and representational capability.

\vspace{-0.3em}
\section{Method}
\vspace{-0.3em}

Our sensitivity-aware visual parameter-efficient fine-tuning consists of two stages. In the first stage, SPT measures the task-specific sensitivity for the pre-trained parameters (Section~\ref{subsec:sensitivity}). Based on the parameter sensitivity and a given parameter budget, SPT then adaptively allocates trainable parameters to task-specific important positions (Section~\ref{subsec:SPT}).

\vspace{-0.3em}
\subsection{Task-specific Parameter Sensitivity}
\label{subsec:sensitivity}
\vspace{-0.3em}

Recent research has observed that pre-trained backbone parameters exhibit varying feature patterns~\cite{raghu2021vision,naseer2021intriguing} and criticality~\cite{zhang2019all,chatterji2019intriguing} at distinct positions. 
Moreover, when transferred to downstream tasks, their efficacy varies depending on how much pre-trained features are reused and how well they adapt to the specific domain gap~\cite{yosinski2014transferable,kumar2022finetuning,neyshabur2020being}. Motivated by these observations, we argue that not all parameters contribute equally to the performance across different tasks in PEFT and propose a new criterion to measure the sensitivity of the parameters in the pre-trained backbone for a given task.

Specifically, given the training dataset $\gD_t$ for the $t$-th task and the pre-trained model weights $\vw=\left\{w_1, w_2, \ldots, w_N\right\}\in \sR^N$ where $N$ is the total number of parameters, the objective for the task is to minimize the empirical risk: $\min_{\vw} E(\gD_t, \vw)$.
We denote the parameter sensitivity \bohan{set} as $\gS=\{s_1, \ldots, s_N\}$ and the sensitivity $s_n$ for parameter $w_n$ is measured by the empirical risk difference when tuning it:
\begin{equation}
\vspace{-0.3em}
    \begin{aligned}
        s_n = E(\gD_t, \vw)-E(\gD_t, \vw\mid w_n=w_n^*),
    \end{aligned}
\label{eq:sensitivity}
\end{equation}
where $w_n^*=\underset{w_n}{\rm argmin}(E(\gD_t, \vw))$. We can reparameterize the tuned parameters as  $w_n^*=w_n+\Delta_{w_n}$, where $\Delta_{w_n}$ denotes the update for $w_n$ after tuning. Here we individually measure the sensitivity of each parameter, which is reasonable given that most of the parameters are frozen during fine-tuning in PEFT. However, it is still computationally intensive to compute Eq.~(\ref{eq:sensitivity}) for two reasons. Firstly, getting the empirical risk for $N$ parameters requires forwarding the entire network $N$ times, which is time-consuming. Secondly, it is challenging to derive $\Delta_{w_n}$, as we have to tune each individual $w_n$ until convergence.

{\begin{algorithm}[t!]
\caption{\label{alg:tps} Computing task-specific parameter sensitivities}
\begin{algorithmic}
    \STATE \textbf{Input:} Pre-trained model with network parameters $\vw$, training set $\gD_t$ for the $t$-th task, and number of training samples $C$ used to calculate the parameter sensitivities
    \STATE \textbf{Output:} Sensitivity set $\gS=\{s_1, \ldots, s_N\}$
    \STATE Initialize $\gS=\{0\}^N$
    \FOR{$i\in\{1,\ldots,C\}$}
        \STATE Get the $i$-th training sample of $\gD_t$
	    \STATE Compute loss $E$
		\STATE Compute gradients $\vg$
		\FOR{$n\in\{1,\ldots,N\}$}
                \STATE Update sensitivity for the $n$-th parameter: $s_{n} = s_{n} + g_n^2$
		    \ENDFOR
    \ENDFOR
\end{algorithmic}
\end{algorithm}}

\begin{figure*}[t]
\begin{center}
    \includegraphics[width=\linewidth]{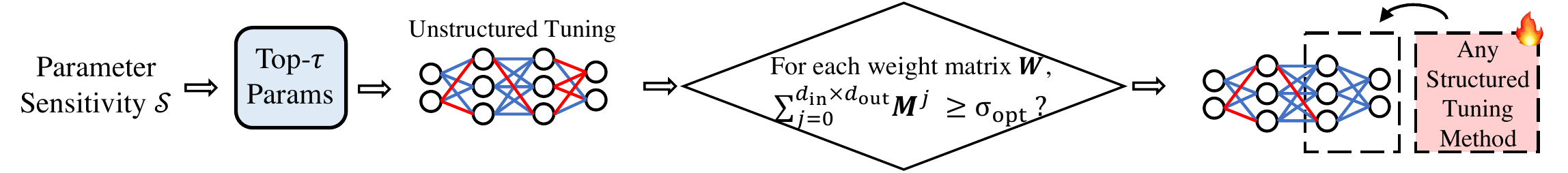}
\end{center}\vspace{-2em}
\caption{Overview of our trainable parameter allocation strategy. With the parameter sensitivity \bohan{set} $\gS$, we first get the top-$\tau$ sensitive parameters. Instead of directly tuning these sensitive parameters, we also boost the representational capability by replacing unstructured tuning with structured tuning at sensitive weight matrices that have a large number of sensitive parameters, which can be implemented by an existing structured tuning method, \eg, LoRA~\cite{hu2022lora} and Adapter~\cite{houlsby2019parameter}. Red lines and blocks represent trainable parameters and modules, while blue lines represent frozen parameters.}
\label{fig:main}
\vspace{-1.5em}
\end{figure*}

To overcome the first barrier, we simplify the empirical loss by approximating $s_n$ in the vicinity of $\vw$ by its first-order Taylor expansion
\vspace{-0.3em}
\begin{equation}
\vspace{-0.5em}
    \begin{aligned}
        s_n^{(1)} = -g_n\Delta_{w_n},
    \end{aligned}
\label{eq:first-order}
\end{equation}
where the gradients $\vg=\partial E/\partial\vw$, and $g_n$ is the gradient of the $n$-th element of $\vg$. 
To address the second barrier, following~\cite{liu2018darts,cai2018proxylessnas}, we take the one-step unrolled weight as the surrogate for $w_n^*$ and approximate $\Delta_{w_n}$ in Eq.~(\ref{eq:first-order}) with a single step of gradient descent. We can accordingly get $s_n^{(1)} \approx g_n^2\epsilon$,
where $\epsilon$ is the learning rate. Since $\epsilon$ is the same for all parameters, we can eliminate it when comparing the sensitivity with the other parameters and finally get 
\vspace{-0.5em}
\begin{equation}
\vspace{-0.3em}
    \begin{aligned}
        s_n^{(1)} \approx g_n^2.
    \end{aligned}
\label{eq:first-order-simp}
\end{equation}
Therefore, the sensitivity of a parameter can be efficiently measured by its potential to reduce the loss on the target domain. Note that although our criterion draws inspiration from pruning work~\cite{molchanov2019importance}, it is distinct from it. \cite{molchanov2019importance} measures the parameter importance by the squared change in loss when removing them, \ie, $\left( E(\gD_t, \vw)-E(\gD_t, \vw\mid w_n=0) \right)^2$ and finally derives the parameter importance by $\left( g_n w_n \right)^2$, which is different from our formulations in Eqs.~(\ref{eq:sensitivity}) and~(\ref{eq:first-order-simp}).

In practice, we accumulate $\gS$ from a total number of $C$ training samples ahead of fine-tuning to generate accurate sensitivity as shown in Algorithm~\ref{alg:tps}, where $C$ is a pre-defined hyper-parameter. In Section~\ref{subsec:abl}, we show that employing only 400 training samples is sufficient for getting reasonable parameter sensitivity, which requires only 5.5 seconds with a single GPU for any VTAB-1k dataset with ViT-B/16 backbone~\cite{vit}.

\vspace{-0.3em}
\subsection{Adaptive Trainable Parameters Allocation}
\label{subsec:SPT}
\vspace{-0.2em}

Our next step is to allocate trainable parameters based on the obtained parameter sensitivity set $\gS$ and a desired parameter budget $\tau$. A straightforward solution is to directly tune the top-$\tau$ most sensitive unstructured connections (parameters) \rev{while keeping the rest frozen}, which we name unstructured tuning. Specifically, we select the top-$\tau$ most sensitive weight connections in $\gS$ to form the sensitive weight connection set $\gT$. Then, for \rev{a} weight matrix $\mW\in \sR^{d_{\rm in}\times d_{\rm out}}$, we can get a binary mask $\mM\in \sR^{d_{\rm in}\times d_{\rm out}}$ computed by
\vspace{-0.5em}
\begin{equation}
\vspace{-0.5em}
    {\begin{array}{ll}
    \small
    \begin{aligned}
    \mM^j =
    \left\{\begin{array}{ll} 
    1 ~~~~~ \mW^j \in \gT \\
    0 ~~~~~ \mW^j \notin \gT
    \end{array}\right.
    \end{aligned},
    \small
    \end{array}}
\label{eq:mask}
\end{equation}
where $\mW^j$ and $\mM^j$ are the $j$-th element in $\mW$ and $\mM$, respectively. Accordingly, we can train the sensitive parameters by gradient descent and the updated weight matrix can be formulated as $\mW'\leftarrow \mW - \epsilon\vg_{\mW}\odot\mM$, where $\vg_{\mW}$ is the gradient for $\mW$.

However, considering PEFT approaches generally limit the proportion of trainable parameters to less than 1\%, tuning only a small number of unstructured weight connections might not have enough representational capability to handle the downstream datasets with large domain gaps from the source pre-training data. Therefore, to improve the representational capability, we propose to replace unstructured tuning with structured tuning at the sensitive weight matrices that have a high number of sensitive parameters. To preserve the parameter budget, we can implement structured tuning with an existing efficient structured tuning PEFT method~\cite{hu2022lora,chen2022adaptformer,houlsby2019parameter,jie2022convolutional} that learns to directly adjust \rev{all hidden dimensions at once}. We depict an overview of our trainable parameter allocation strategy in Figure~\ref{fig:main}. For example, we can employ the low-rank reparameterization trick LoRA~\cite{hu2022lora} to the sensitive weight matrices \rev{and the one-step update for $\mW$ can be formulated as}
\vspace{-0.4em}
\begin{equation}
\vspace{-0.4em}
    {\begin{array}{ll}
    \small
    \begin{aligned}
    \mW' = \left\{\begin{array}{ll} 
    \mW + \mW_{\rm down}\mW_{\rm up} & ~~ \text { if } ~~ \sum_{j=0}^{d_{\rm in}\times d_{\rm out}} \mM^j \geq \sigma_{\rm opt} \\
    \mW - \epsilon\vg_{\mW}\odot\mM & ~~ {\rm otherwise}
    \end{array}\right.
    \end{aligned},
    \small
    \end{array}}
\label{eq:weight_updat}
\end{equation}
where $\mW_{\rm down}\in \sR^{d_{\rm in}\times r}$ and $\mW_{\rm up}\in \sR^{r\times d_{\rm out}}$ are two learnable low-rank matrices to approximate the update of $\mW$ and rank $r$ is a hyper-parameter where $r \ll {\rm min}(d_{\rm in},d_{\rm out})$. In this way, we perform structured tuning on $\mW$ when its number of sensitive parameters exceeds $\sigma_{\rm opt}$, whose value depends on the pre-defined type of structured tuning method. For example, since implementing structured tuning with LoRA requires $2\times d_{\rm in} \times d_{\rm out} \times r$ trainable parameters for each sensitive weight matrix, we set $\sigma_{\rm LoRA} \leftarrow 2\times d_{\rm in} \times d_{\rm out} \times r$ to ensure that the number of trainable parameters introduced by structured tuning is always equal to or lower than the number of sensitive parameters.

In this way, our SPT adaptively incorporates both structured and unstructured tuning granularities to enable higher flexibility and stronger representational power, simultaneously. In Section~\ref{subsec:abl}, we show that structured tuning is important for the downstream tasks with larger domain gaps and both unstructured and structured tuning contribute clearly to the superior performance of our SPT.

\begin{table*}[t]
\centering
\resizebox{0.8\linewidth}{!}{
\begin{tabular}{
l  !{\color{tabvline}\vrule}
r  !{\color{tabvline}\vrule}
rr   !{\color{tabvline}\vrule}
rrrrr
}
\toprule
\textbf{\vit{}-B/16} & \textbf{Total}
&\multicolumn{2}{c!{\color{tabvline}\vrule}}{\bf{FGVC}}
&\multicolumn{5}{c}{\bf{\vtab{}}}
\\
\bf{(85.8M)} & \textbf{params}
&\bf{\scriptsize{Tuned / Total}}
&\bf{\scriptsize{Mean Acc.}}
&\bf{\scriptsize{Tuned / Total}}
&\bf{\scriptsize{Natural}} &\bf{\scriptsize{Specialized}} &\bf{\scriptsize{Structured}}
&\bf{\scriptsize{Mean Acc.}}
\\
\midrule
\band \fullft{} & 24.02$\times$ & 100\%
&88.5 & 100\% &75.9 & 83.4 & 47.6& 69.0
\\\midrule
 \multicolumn{9}{c}{\bf{Addition-based methods}}
\\\midrule
\mlp-3 & 1.35$\times$ &1.50\%
&79.8 & 1.42\% &67.8 &72.8 &30.6 & 57.1
\\
\shallowprompt{} & 1.04$\times$ & 0.31\%
&84.6 & 0.13\% &76.8 &79.7 & 47.0 & 67.8
\\\deepprompt{} & 1.18$\times$ &  0.98\%
& 89.1 & 1.14\%
&78.5 &82.4 &55.0 & 72.0
\\
\adapter{}-8 & 1.06$\times$ & 0.39\%
& 85.5 & 0.23\% &79.0 & 84.1 & 58.5 & 73.9
\\
\adapter{}-32 & 1.19$\times$ & 0.95\%
&85.6 & 0.71\% &79.6 &84.0 &58.3 & 74.0
\\
\adaptformer{} & 1.09$\times$ & 0.44\% &
85.1 & 0.36\%
& 80.6 & \underline{85.4} & 58.5 & 74.8 \\
\noah{} & - & - &
- & 0.52\%
& 80.2 & 84.9 & \underline{61.3} & 75.5 \\
\SPTa{} (Ours) & 1.08$\times$ & 0.41\%
& \underline{89.5} & 0.30\% & \underline{81.3} & 85.3 & 60.8 & \underline{75.8}\\
\SPTa{} (Ours) & 1.10$\times$ & 0.47\%
& \textbf{89.8} & 0.44\% & \textbf{82.0} & \textbf{85.8} & \textbf{61.4} & \textbf{76.4}
\\
\midrule
 \multicolumn{9}{c}{\bf{Reparameterization-based methods}}
\\\midrule
\linear{} & 1.02$\times$ & 0.12\%
&79.3 & 0.04\% &68.9 &77.2 &26.8 & 57.6
\\
\partialft{}-1 & 3.00$\times$ &8.38\%
&82.6 & 8.30\% &69.4 &78.5 &34.2 & 60.7
\\
\bias{} & 1.05$\times$ & 0.13\%
&88.4 & 0.13\% &73.3 &78.3 &44.1 & 65.2
\\
\lora{}-8 & 1.07$\times$ & 0.55\%
& 86.0 & 0.23\% &79.5 & 84.6 & 60.5 & 74.9
\\
\lora{}-16 & 1.18$\times$ & 0.90\%
& 84.8 & 0.69\% &79.8 &84.9 & 60.2 & 75.0
\\
\SPTl{} (Ours) & 1.08$\times$ & 0.41\%
& \underline{89.3} & 0.31\% & \underline{81.5} & \underline{85.6} & \underline{60.7} & \underline{75.9} \\
\SPTl{} (Ours) & 1.15$\times$ & 0.60\%
& \textbf{90.1} & 0.63\% & \textbf{81.9} & \textbf{85.9} & \textbf{61.3} & \textbf{76.4} \\
\bottomrule
\end{tabular}
}
\vspace{-0.5em}
\caption{Comparisons on FGVC and \vtab{}~\cite{zhai2019vtab} benchmarks using supervised pre-trained ViT-B/16 backbone pre-trained on ImageNet-21k. ``Total params'' denotes the ratio of the total number of parameters needed for all downstream tasks relative to the one for the pre-trained backbone, and ``Tuned/Total'' denotes the fraction of trainable parameters. Top-1 accuracy (\%) is reported. The best result is in \textbf{bold}, and the second-best result is \underline{underlined}.}
\label{tab:main_sup}
\vspace{-0.5em}
\end{table*}

\vspace{-0.2em}
\section{Experiments}
\vspace{-0.2em}
\subsection{Experimental Setup}
\vspace{-0.2em}

\noindent\textbf{Datasets and metrics.} We evaluate our SPT on total $24$ downstream tasks in two groups following~\cite{jia2022vpt}. 1) FGVC is a benchmark for fine-grained visual classification, including \cub{}~\cite{wah2011caltech}, \nabirds{}~\cite{van2015building}, \flowers{}~\cite{nilsback2008automated}, \cars{}~\cite{gebru2017cars}, and \dogs{}~\cite{Khosla_FGVC2011dogs} datasets. Each FGVC dataset contains between 55 to 200 classes and a few thousand images for train, validation, and test. We follow the validation splits in~\cite{jia2022vpt} if the validation set is unavailable. 2) \vtab{}~\cite{zhai2019vtab} is a large-scale transfer learning benchmark consisting of a collection of 19 visual classification tasks. \vtab{} can further be divided into three groups, including Natural tasks with natural images, Specialized tasks with images captured by specialized equipment, e.g., medical images, and Structured tasks with images mostly generated from synthetic environments. Each of the \vtab{} dataset has only 800 training and 200 validation samples, while the test set sizes vary. We use top-1 accuracy (\%) averaged within each group as our main metric following~\cite{jia2022vpt}.

\noindent\textbf{Pre-trained backbones.} We conduct experiments on the plain vision Transformer backbone ViT-B/16~\cite{vit} that is pre-trained on \imagenet{}~\cite{krizhevsky2012imagenet} with different pre-training strategies following~\cite{jia2022vpt}, including supervised pre-training and self-supervised pre-training with \mae{}~\cite{he2022masked} and \moco{}~\cite{chen2021empirical} following~\cite{jia2022vpt}. We also conduct experiments on the representative hierarchical vision Transformer backbone Swin-B~\cite{swin} under supervised pre-training.

\noindent\textbf{Contenders.} We categorize the baseline methods into addition-based and reparameterization-based PEFT methods as introduced in Section~\ref{subsec:related}. Unless specified, all baseline methods keep the backbone frozen. Addition-based methods require extra computations during inference, including \mlp{}-$k$, \shallowprompt{}~\cite{jia2022vpt}, \deepprompt{}~\cite{jia2022vpt}, \adapter{}-$k$~\cite{houlsby2019parameter}, \adaptformer{}~\cite{chen2022adaptformer}, and \noah{}~\cite{zhang2022neural}. Reparameterization-based methods have no additional computational overhead during inference, including \linear{}, \partialft{}-$k$, \bias{}~\cite{zaken2022bitfit}, and \lora{}-$k$~\cite{hu2022lora}. Here $k$ represents the number of bottleneck dimension in \adapter{}-$k$ and \lora{}-$k$. We also 
compare with full fine-tuning which is denoted by \fullft{}. We introduce the details of these methods in the supplementary material.

We also introduce two variants of our SPT: addition-based \SPTa{} and reparameterization-based \SPTl{}. \SPTa{} directly adjusts the hidden representations that are computed by sensitive weight matrices following~\cite{houlsby2019parameter}, while \SPTl{} approximates updating the sensitive weight matrices following~\cite{hu2022lora}. For the two variants, we follow the exact weight initializations that are described in~\cite{hu2022lora} and follow~\cite{zhang2022neural} to set the bottleneck dimension as 8.

\noindent\textbf{Implementation details.}
Following~\cite{zhang2022neural}, we use the AdamW optimizer~\cite{loshchilov2018fixing} with cosine learning rate decay and set the batch size, learning rate, and weight decay as 64, $1\times10^{-3}$, and $1\times10^{-4}$, respectively. We also follow~\cite{zhang2022neural} for the standard data augmentation pipeline. We set the number of training samples $C$ used to calculate our parameter sensitivities in Algorithm~\ref{alg:tps} universally to be 800 for the main experiments.

\begin{table*}[t]
\centering
\resizebox{\textwidth}{!}{
\begin{tabular}{
l  !{\color{tabvline}\vrule}
r  !{\color{tabvline}\vrule}
rrrrr   !{\color{tabvline}\vrule}
rrrrr}
\toprule
\textbf{\vit{}-B/16}
&\bf{\scriptsize{Total}}
&\multicolumn{5}{c!{\color{tabvline}\vrule}}{  \bf{\vtab{}} \quad \mae{}}
&\multicolumn{5}{c}{\bf{\vtab{}} \quad  \moco{}}
\\
\bf{(85.8M)}
&\bf{\scriptsize{Params}}
&\bf{\scriptsize{Tuned / Total}}
&\bf{\scriptsize{Natural}} &\bf{\scriptsize{Specialized}} &\bf{\scriptsize{Structured}}
&\bf{\scriptsize{Mean Acc.}}
&\bf{\scriptsize{Tuned / Total}}
&\bf{\scriptsize{Natural}} &\bf{\scriptsize{Specialized}} &\bf{\scriptsize{Structured}}
&\bf{\scriptsize{Mean Acc.}}
\\
\midrule
\band \fullft{} & 38.02$\times$ & 100\%
&59.3 &79.7 &53.8 &64.3 & 100\% &72.0 &84.7 &42.0 & 69.6
\\\midrule
 \multicolumn{12}{c}{\bf{Addition-based methods}}
\\\midrule
\adapter{}-8 &1.08$\times$ & 0.23\%
& 57.2 & 78.4 & 54.7 & 63.4 & 0.23\% & 27.6 & 70.9 & 48.4 & 49.0 
\\
\adapter{}-32 &1.28$\times$ & 0.95\%
& 55.3 & 78.8 & 53.3 & 62.5 & 0.99\% & 74.2 & 82.7 & 47.7 & 68.2
\\
\shallowprompt{} & 1.02$\times$ & 0.12\%
& 40.0 & 69.7 & 27.5 & 45.7 & 0.12\%  & 67.3 & 82.3 & 37.6 & 62.4
\\\deepprompt{} & 1.05$\times$ & 0.23\%  & 36.0 & 60.6 & 26.6
& 41.1 & 0.07\%
&70.3 & 83.0 & 42.4 & 65.2\\
\SPTa{} (Ours) & 1.07$\times$ & 0.26\% & \underline{64.8} & \underline{82.4} & \underline{60.4} & \underline{69.2} & 0.08\% & \underline{76.1} & \underline{84.9} & \underline{60.1} & \underline{73.7}
\\
\SPTa{} (Ours) & 1.13$\times$ & 0.41\% & \textbf{65.6} & \textbf{82.7} & \textbf{60.7} & \textbf{69.7} & 0.30\% & \textbf{76.6} & \textbf{85.0} & \textbf{61.7} & \textbf{74.4}
\\\midrule
 \multicolumn{12}{c}{\bf{Reparameterization-based methods}}
\\\midrule
\linear{} & 1.02$\times$ & 0.04\% & 18.9 & 52.7 & 23.7 & 32.1 & 0.04\% & 67.5 &81.1 & 30.3 & 59.6
\\
\partialft{}-1 & 4.16$\times$ &8.30\%
&58.4 & 78.3 & 47.6 & 61.5 & 8.30\% &72.3 & 84.6 &47.9 & 68.3
\\
\bias{} & 1.06$\times$ & 0.13\%
&54.6 & 75.7 & 47.7 & 59.3 & 0.13\% & 72.9 &81.1 &53.4 & 69.2
\\
\lora{}-8 &1.08$\times$ & 0.23\%
& 57.5 & 77.7 & 57.7 & 64.3 & 0.23\% & 21.2 & 66.7 & 45.1 & 44.3
\\
\lora{}-16 &1.28$\times$ & 0.69\%
& 57.3 & 77.1 & 59.9 & 64.8 & 0.69\% & 16.0 & 64.0 & 48.7 & 42.9
\\
\SPTl{} (Ours) & 1.11$\times$& 0.29\% & \underline{63.8} & \underline{81.6} & \underline{60.0} & \underline{68.5} & 0.30\% & \textbf{76.5} & \underline{85.4} & \underline{63.0} & \underline{75.0} \\
\SPTl{} (Ours) & 1.23$\times$& 0.69\% & \textbf{65.4} & \textbf{82.4} & \textbf{61.5} & \textbf{69.8} & 0.50\% & \textbf{76.5} & \textbf{86.0} & \textbf{63.6} & \textbf{75.3} \\
\bottomrule
\end{tabular}
}
\vspace{-0.5em}
\caption{Comparisons on \vtab{}~\cite{zhai2019vtab} benchmark using self-supervised ViT-B/16 backbone pre-trained by \mae~\cite{he2022masked} and \moco~\cite{chen2021empirical}. ``Total params'' denotes the ratio of the total number of parameters needed for all downstream tasks relative to the one for the pre-trained backbone, and ``Tuned/Total'' denotes the fraction of trainable parameters. Top-1 accuracy (\%) is reported. The best result is in \textbf{bold}, and the second-best result is \underline{underlined}.}
\label{tab:main_ssup}
\vspace{-1.5em}
\end{table*}

\begin{table}[t!]
\centering
\resizebox{\linewidth}{!}{%
    \begin{tabular}{l!{\color{tabvline}\vrule}ccccc}
    \toprule
    \bf{Method} & \begin{tabular}[c]{@{}c@{}} \bf{Tuned / } \\  \bf{Total} \end{tabular}  &\bf{Natural} &\bf{Specialized} &\bf{Structured}
    & \begin{tabular}[c]{@{}c@{}} \bf{Mean / } \\  \bf{Acc.} \end{tabular} \\
    \midrule
    \band \fullft{} & 100\% & 79.1 & 86.2 & 59.7 & 75.0
    \\\midrule
     \multicolumn{6}{c}{\bf{Addition-based methods}}
    \\\midrule
    \mlp{}-3 & 1.60\% & 73.6 & 75.2 & 35.7 & 61.5 \\
    \shallowprompt{} & 0.04\% & 79.9 & 82.5 & 37.8 & 66.7  \\
    \deepprompt{} & 0.23\% & 76.8 & 84.5 & 53.4 & 71.6 \\
    \adapter{}-8 & 1.18\% & 81.7 & \textbf{87.3} & 61.2 & 76.7 \\
    \SPTa{} (ours) & 0.33\% & \textbf{83.0} & \textbf{87.3} & \textbf{62.1} & \textbf{77.5}
    \\\midrule
     \multicolumn{6}{c}{\bf{Reparameterization-based methods}}
    \\\midrule
    \linear{} & 0.04\% & 73.5 & 80.8 & 33.5 & 62.6 \\
    \partialft{}-1 & 2.15\% & 73.1 & 81.7 & 35.0 & 63.3 \\
    \lora{}-8 & 1.18\% & 81.7 & 87.2 & 60.1 & 76.3 \\
    \SPTl{} (ours) & 0.49\% & \textbf{83.1} & \textbf{87.4} & \textbf{60.4} &  \textbf{77.2} \\
    \bottomrule
    \end{tabular}%
    }
        \vspace{-0.5em}
    \caption{Comparisons on \vtab{}~\cite{zhai2019vtab} benchmark with supervised pre-trained Swin-B~\cite{swin}. ``Tuned/Total'' denotes the fraction of trainable parameters. Top-1 accuracy (\%) is reported. The best result is in \textbf{bold}.}
        \vspace{-1.5em}
    \label{tab:main_swin}
\end{table}

\vspace{-0.3em}
\subsection{Main Results}
\vspace{-0.2em}
\label{subsec:main_results}

We compare our method with the baseline methods under different backbones, pre-training strategies, and tasks.

\begin{figure*}[t]
\begin{center}
\includegraphics[width=\linewidth]{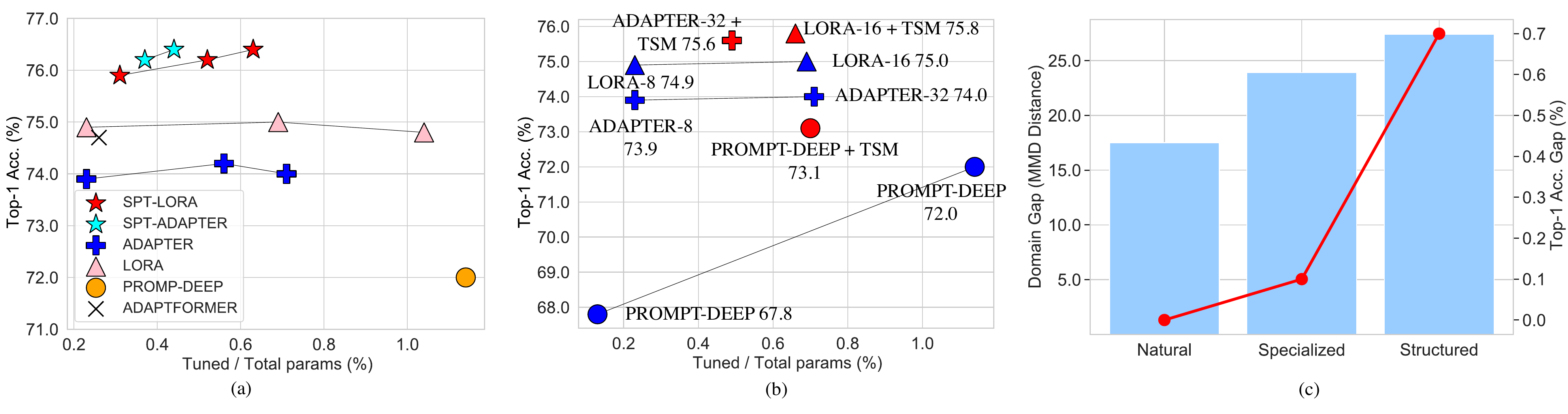}
\end{center}
\vspace{-2em}
\caption{ (a) Accuracy vs. parameter efficiency with supervised pre-trained ViT-B/16 backbone on \vtab{}~\cite{zhai2019vtab}. SPT variants perform favorably against the other PEFT approaches and are more scalable. (b) Applying other PEFT structured tuning methods~\cite{jia2022vpt,houlsby2019parameter,hu2022lora} to the task-specific sensitive weight matrices (denoted by TSM) identified by our criterion with supervised pre-trained ViT-B/16 backbone on \vtab{}. Our criterion brings consistent performance gain. (c) Domain vs. performance gaps for different dataset groups in VTAB-1k~\cite{zhai2019vtab}. The blue bars show the domain gaps between the source domain (\imagenet{}~\cite{krizhevsky2012imagenet}) and target domains, which are measured by Maximum Mean Discrepancy (MMD) distance~\cite{tzeng2014deep}. The red line represents the performance gaps between \SPTa{} w/o unstructured and w/o structured, using supervised pre-trained ViT-B/16 backbone. The dataset groups are Natural, Specialized, and Structured. Structured tuning is important for achieving good performance on Structured datasets with larger domain gaps.}
\vspace{-1.em}
\label{fig:abl}
\end{figure*}

\noindent\textbf{Visual recognition with ViT backbone.} First, \emph{our proposed \SPTa{} and \SPTl{} achieve the best performance under different trainable parameter budgets} with supervised pre-trained ViT-B/16 backbone, as shown in Table~\ref{tab:main_sup} and Figure~\ref{fig:abl}~(a). For instance, \SPTa{} outperforms the SOTA method \noah{} by a clear margin of 0.9\% mean top-1 accuracy over the 19 \vtab{} datasets with fewer trainable parameters. We speculate that our SPT variants allocate trainable parameters at task-specific positions compared to the heuristically selected positions in the baseline methods, which contributes to our superior performance. We also observe that our \SPTa{} and \SPTl{} achieve large performance gains over \adapter{} and \lora{} variants, respectively. For example, \SPTa{} and \SPTl{} with 0.41\% trainable parameters respectively improve \adapter{}-8 and \lora{}-8 significantly by 4.0\% and 3.3\% mean accuracy on the FGVC benchmark. This suggests that identifying task-specific important positions and combining both unstructured and structured tuning granularities with SPT are complementary to the existing PEFT methods and boost their performance.

Second, \emph{\SPT{} variants outperform baseline methods and full fine-tuning by significant margins with the self-supervised pre-trained ViT-B/16 backbones.} As shown in Table~\ref{tab:main_ssup}, existing PEFT approaches exhibit inferior results than full fine-tuning with the self-supervised pre-trained backbones \mae{} and \moco{}.
It is worth noting that
previous PEFT methods yield inconsistent results with the backbones of different pre-training strategies. In contrast,
SPT variants consistently outperform full fine-tuning. In particular, \SPTa{} achieves remarkable 5.8\% and 5.5\% mean top-1 accuracy gains over the best-performing baseline method on \vtab{} benchmark with only 0.26\% and 0.08\% trainable parameters for \mae{} and \moco{} pre-trained backbones, respectively. Moreover, our observation in supplementary material suggests that self-supervised pre-trained ViT backbones have more diverse sensitivity distributions and a higher variance in sensitivity across different tasks than the supervised pre-trained one. This leads to the conjecture that baseline methods which assign trainable parameters to the same positions for all tasks may fail to mitigate the distinct domain gaps in individual downstream datasets, whereas our SPT allocates trainable parameters to task-specific positions accurately. 

\noindent\textbf{Visual recognition with Swin and ConvNeXt backbones.}
From Table~\ref{tab:main_swin}, we observe that our \SPTl{} and \SPTa{} also achieve SOTA performance with Swin-B backbone on all dataset groups. We also follow VPT~\cite{jia2022vpt} to apply SPT variants to ResNet-alike architecture ConvNeXt-Base~\cite{liu2022convnet} and report the results in Table~\ref{tab:res}. We observe that SPT variants achieve better trade-offs between accuracy and parameter efficiency than the baseline methods. The results further demonstrate the versatility and effectiveness of our SPT.

\noindent\textbf{Semantic segmentation.} We follow VPT~\cite{jia2022vpt} to conduct semantic segmentation on ADE20k dataset. Following the settings of~\cite{jia2022vpt}, we apply our SPT to SETR-PUP~\cite{zheng2021rethinking} framework with ViT-L backbone and only allocate trainable parameters to the backbone with the head fully fine-tuned. We report the mIoU results in Table~\ref{tab:seg}. Notably, \SPTl{} and \SPTa{} outperform the baseline methods by large margins, indicating our SPT can be generalized to the semantic segmentation task. We report the results for conducting only structured tuning for both \SPTl{} and \SPTa{} as it yields higher mIoU.

\begin{table}[tb!]
\centering
    \resizebox{\linewidth}{!}{%
    \rev{
    \begin{tabular}{lc!{\color{tabvline}\vrule}cccc}
    \toprule
    \bf{Method} & \bf{Tuned/Total} & \bf{Natural} & \bf{Specialized} & \bf{Structured} & \bf{Mean Acc.} \\\midrule
    FULL & 100\% & 78.0 & 83.7 & 60.4 & 74.0 \\
    LoRA & 0.79\% & 82.2 & 84.7 & 64.1 & 77.0 \\
    Adapter & 0.47\% & 83.1 & 84.9 & 64.6 & 77.5 \\\midrule
    SPT-LoRA (ours) & 0.57\%  & 83.4 & \textbf{86.7} & \textbf{65.9} & \textbf{78.7} \\
    SPT-Adapter (ours) & 0.36\% & \textbf{83.7}
     & 86.2 & 65.3 & 78.4\\\bottomrule
    \end{tabular}%
    }}
    \caption{Comparisons on \vtab{}~\cite{zhai2019vtab} benchmark with supervised pre-trained ConvNeXt-Base~\cite{liu2022convnet} backbone. ``Tuned/Total'' denotes the fraction of trainable parameters. Top-1 accuracy (\%) is reported. The best result is in \textbf{bold}.
}\label{tab:res}
\end{table}

\begin{table}[tb!]
\centering
    \resizebox{\linewidth}{!}{%
    \rev{
    \begin{tabular}{l!{\color{tabvline}\vrule}ccc}
    \toprule
    \bf{Method} & \bf{mIoU-s.s. (\%)} & \bf{mIoU-m.s. (\%)} & \bf{Trainable Param.} \\ \midrule
    VPT & 44.0 & 45.6 & 15.8M  \\
    LoRA & 43.9 & 45.9 & 14.7M \\
    Adapter & 44.4 & 46.6 & 14.6M \\\midrule
    SPT-LoRA (ours) & \textbf{45.4} & \textbf{47.5} & 14.6M \\
    SPT-Adapter (ours) & 45.2 & 47.2 & 14.6M \\\bottomrule
    \end{tabular}%
    }}
    \caption{Semantic Segmentation: Comparisons on ADE20k~\cite{zhou2017scene} \texttt{val} with SETR~\cite{zheng2021rethinking} on ViT-L backbone. We report both single-scale (s.s.) and multi-scale (m.s.) mIoU results. The best result is in \textbf{bold}.
}\label{tab:seg}
\end{table}

\vspace{-0.2em}
\subsection{Ablation Study}
\label{subsec:abl}
\vspace{-0.2em}
\noindent\textbf{Effect of the sensitivity criterion.} We investigate the effectiveness of our sensitivity criterion on~\vtab{} by employing structured tuning methods from~\cite{jia2022vpt,houlsby2019parameter,hu2022lora} to the task-specific sensitive weight matrices. Note that we do not conduct unstructured tuning to ensure fair comparisons. The results are presented in Figure~\ref{fig:abl}~(b). Our criterion brings consistent 1.1\%, 1.6\%, and 0.8\% performance gains for \deepprompt{}, \adapter{}-32, and \lora{}-16, respectively, which demonstrates the effectiveness and versatility of our sensitivity criterion to identify accurate task-specific important positions.

\noindent\textbf{Effect of structured and unstructured tuning.} We investigate the effectiveness of unstructured and structured tuning individually on~\vtab{}. The results are presented in Table~\ref{tab:structured}. We start by applying \adapter{}-8 to the sensitive weight matrices identified by our sensitivity criterion (\SPTa{} w/o unstructured). We observe that our sensitivity criterion boosts the performance of all the dataset groups by clear margins, which again demonstrates the importance of our sensitivity criterion. Next, we observe that allocating the trainable parameters to the unstructured sensitive weight connections also brings accuracy improvement to the Natural and Specialized datasets from \adapter{}-8. However, we find that structured tuning is especially important for achieving good performance on Structured datasets. To further investigate this phenomenon, we observe that Structured datasets have larger domain gaps from the pre-training source domain~\cite{krizhevsky2012imagenet} compared to Natural and Specialized datasets as visualized in Figure~\ref{fig:abl}~(c). We hence conjecture that structured tuning has a higher representational capability than unstructured tuning which facilitates mitigating the large domain gaps during fine-tuning (see the supplementary material for visual examples). Finally, we observe that incorporating both structured and unstructured tuning at task-specific important positions achieves the highest performance on all dataset groups.

\noindent\textbf{Effect of number of training samples $C$ to get parameter sensitivity.} We investigate the effect of the number of training images $C$ for calculating our parameter sensitivity (Algorithm~1 of the main paper). We randomly sample training samples and report the mean results over three runs in Table~\ref{tab:num_samples}. We find that our SPT is robust to the number of training samples $C$ and randomly sampling 400 out of a total of 800 training samples is sufficient to obtain accurate task-specific important positions, \eg, calculating the sensitivity for ViT-B/16 backbone takes only 5.5 seconds with a single GPU on any of the \vtab{} datasets and this computation is required only once.

\begin{figure*}[!th]
\begin{center}
    \includegraphics[width=\linewidth]{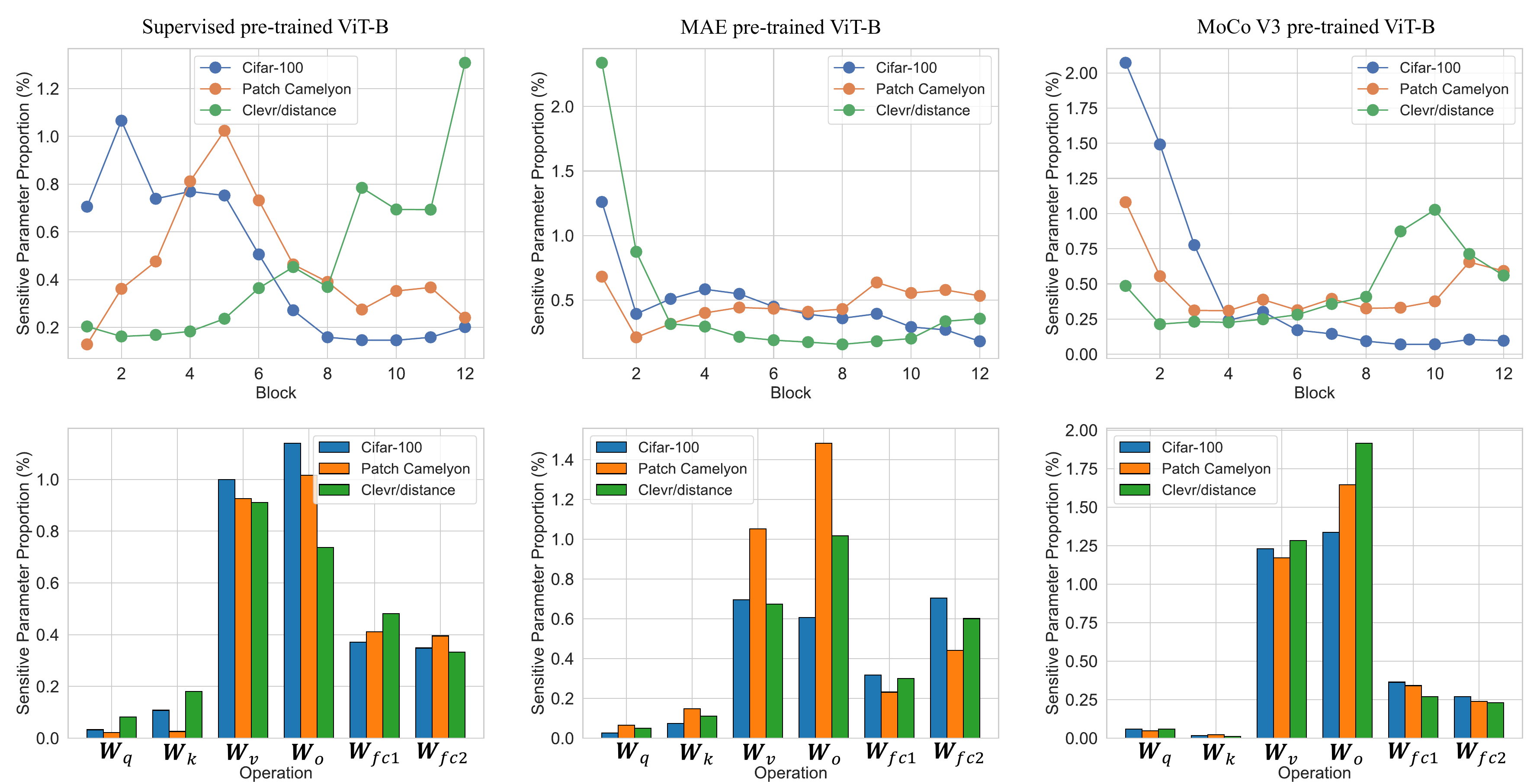}
\end{center}\vspace{-1.5em}
\caption{Parameter sensitivity patterns under 0.4M trainable parameter budget for ViT-B/16 backbone with different pre-training strategies on three sample tasks from VTAB-1k~\cite{zhai2019large}. The proportions exhibit task-specific varying patterns in terms of network depth (upper figures) and task-agnostic similar patterns in terms of operations (lower figures).}
\label{fig:sensitivity_depth}
\vspace{-1em}
\end{figure*}

\begin{table}[t!]
\centering
\resizebox{\linewidth}{!}{%
    \begin{tabular}{l!{\color{tabvline}\vrule}ccccc}
    \toprule
    \bf{Method} & \begin{tabular}[c]{@{}c@{}} \bf{Tuned / } \\  \bf{Total} \end{tabular}  &\bf{Natural} &\bf{Specialized} &\bf{Structured}
    & \begin{tabular}[c]{@{}c@{}} \bf{Mean / } \\  \bf{Acc.} \end{tabular} \\
    \midrule
    \adapter{}-8 & 0.23\% & 79.0 & 84.1 & 58.5 & 73.9 \\
    \SPTa{} w/o unstructured & 0.29\% & 81.2 & 85.1 & 60.3 & 75.5 \\
    \SPTa{} w/o structured & 0.34\% & 81.2 & 85.0 & 59.6 & 75.3 \\
    \SPTa{} & 0.30\% & \textbf{81.3} & \textbf{85.3} & \textbf{60.8} & \textbf{75.8} \\
    \bottomrule
    \end{tabular}%
    }
    \caption{Ablation study on structured and unstructured tuning only with supervised pre-trained ViT-B/16 backbone. Top-1 accuracy (\%) is reported. We set different parameter constraints to align the fractions of trainable parameters for these cases. The best result is in \textbf{bold}}
\vspace{-1.em}
    \label{tab:structured}
\end{table}

\begin{table}[tb!]
\centering
    \resizebox{\linewidth}{!}{%
    \begin{tabular}{l!{\color{tabvline}\vrule}ccc}
    \toprule
     \bf{Method}  &  \bf{\begin{tabular}[c]{@{}c@{}} Inference Latency \\ (ms/img) \end{tabular}  }  & \bf{\begin{tabular}[c]{@{}c@{}} Inference Memory \\ (GB) \end{tabular}} & \bf{\begin{tabular}[c]{@{}c@{}} Fine-tuning Memory \\ (GB) \end{tabular}} \\ 
    \midrule
    \fullft{}  & \textbf{2.8} & \textbf{1.3} & 11.9 \\
    \deepprompt{}  & 3.8 & 1.9 & 13.2 \\
    \lora-16  & \textbf{2.8} & \textbf{1.3} & 8.2 \\
    \SPTl{} w/o unstructured  & \textbf{2.8} & \textbf{1.3} & 8.3 \\
    \SPTl{}  & \textbf{2.8} & \textbf{1.3} & \rev{9.8} \\
    \bottomrule
    \end{tabular}%
    }\vspace{-0.5em}
\caption{Cost comparisons with ViT-B/16 backbone on the \vtab{}~\cite{zhai2019vtab} benchmark with around 0.70\% fractions of the trainable parameters. We report the latency (ms/img) and the peak memory usage (GB) with image resolution 224$\times$224 on a single GeForce 3090 GPU. The best result is in \textbf{bold}.}
\label{tab:comsumption}
\end{table}

\begin{table}[t!]
\centering
    \resizebox{0.6\linewidth}{!}{%
    \begin{tabular}{l!{\color{tabvline}\vrule}cccc}
    \toprule
     $C$ & 240 & 400 & 560 & 800 \\ \midrule
     \bf{\scriptsize{Mean Acc.}} & 76.3 & \textbf{76.4} & \textbf{76.4}& \textbf{76.4}\\
    \bottomrule
    \end{tabular}%
    }
        \vspace{-0.5em}
\caption{Effect of the number of training samples used to get the sensitivity for \SPTl{} with supervised pre-trained ViT-B/16 backbone on \vtab{}~\cite{zhai2019vtab}. Top-1 accuracy (\%) is reported. The best result is in \textbf{bold}.
}\vspace{-1.0em}
\label{tab:num_samples}
\end{table}

\noindent\textbf{Computational cost analysis.} We investigate the computational cost of \SPTl{} by comparing with full fine-tuning, addition-based method \deepprompt{}, and reparameterization-based method \lora{}-16. 
The results are presented in Table~\ref{tab:comsumption}. We observe that \deepprompt{} has higher inference latency and inference GPU memory due to the additional prompts. In contrast, since the updated parameters after fine-tuning can be reparameterized and merged into the pre-trained model, our \SPTl{}, \SPTl{} w/o unstructured tuning, and \lora{}-16 are more efficient than \deepprompt{} during inference. 
However, we observe that our \SPTl{} has slightly higher fine-tuning memory than the full fine-tuning and \lora{}-16 which is taken up by \rev{updating the unstructually-tuned parameters with sparse gradients} in Eq.~(\ref{eq:weight_updat}). Additionally, for memory-intense scenarios, one can employ SPT-LoRA w/o unstructured tuning for improved performance (0.8\% Top-1 accuracy on VTAB-1k) and similar fine-tuning memory as LoRA, as shown in Figure~\ref{fig:abl}~(b) and Table~\ref{tab:structured}.

\vspace{-0.3em}
\subsection{Observations on Sensitivity Patterns}\label{subsec:observation}
\vspace{-0.3em}

Our sensitivity criterion identifies task-specific important positions, which can reveal the contributions of the pre-trained weights to different downstream tasks during transfer learning. We visualize the proportions of the sensitive parameters for the supervised pre-trained ViT-B/16 backbone under 0.4M trainable parameter budget in Figure~\ref{fig:sensitivity_depth}. First, we investigate the most sensitive blocks, whose numbers of sensitive parameters are summed and normalized over the 12 ViT-B/16 blocks. We observe that the patterns of the sensitive parameter proportions vary markedly across different tasks, which echoes the observations made in~\cite{guo2019spottune}. This suggests that we should not introduce trainable parameters to the same positions for each individual task but allocate trainable parameters at task-specific ones as we proposed. Next, we investigate the most insensitive weight matrices within a block. A ViT block consists of a query $\mW_{q}$, a key $\mW_{k}$, a value $\mW_{v}$, and an output $\mW_{o}$ weight matrices in the multi-head self-attention layer and two weight matrices $\mW_{fc1}$ and $\mW_{fc2}$ in the feed-forward network as elaborated in~\cite{vaswani2017attention,vit}. We observe that the query $\mW_{q}$ and key $\mW_{k}$ weight matrices have the lowest proportions of sensitive parameters for all three sample tasks. Since $\mW_{q}$ and $\mW_{k}$ are responsible for learning the attention scores which indicate the pairwise similarity among the patches, we speculate that although domain changes, the patch relationships learned during pre-training can be efficiently reused when transferred to downstream classification tasks.

\section{Conclusion}
\vspace{-0.3em}
In this paper, we have explored identifying and allocating trainable parameters to task-specific important positions for visual parameter-efficient tuning. Specifically, we have proposed a novel criterion to quickly measure the sensitivity of the pre-trained parameters for each specific task before fine-tuning. Based on the parameter sensitivity, we have proposed a trainable parameter allocation strategy that adaptively combines both unstructured and structured tuning under a desired trainable parameter budget, enabling high representational capability and flexibility. Finally, we have conducted extensive experiments on a total of 24 downstream recognition tasks with both plain and hierarchical vision Transformer backbones under different pre-training strategies to demonstrate the versatility and effectiveness of our proposed SPT. Notably, we have shown that our approach is complementary to the existing PEFT methods and improves their performance significantly. In the future, we will explore adapting large vision models to more downstream tasks with SPT, \eg, dense prediction and vision-and-language tasks, and improve the training efficiency of SPT for on-device training~\cite{cai2020tinytl,lin2022device}. 

\noindent\textbf{Acknowledgement.} We thank Jing liu and Ziyi Liu for their helpful discussions. This research is partially supported by Monash FIT Start-up Grant. Dr. Jing Zhang is supported by the Australian Research Council project FL-170100117.

\clearpage

\bibliographystyle{abbrv}
{\small
 \bibliography{egbib}
}

\onecolumn
\section*{Appendix}

We organize our supplementary material as follows. 
\begin{itemize}
    \item In Section~\ref{subsec:supp_contenders}, we introduce more details about the contenders.
    \item In Section~\ref{subsec:supp_pattern1}, we show more sensitivity patterns for ViT-B/16 with various pre-training strategies.
    \item In Section~\ref{subsec:supp_visual}, we show some dataset samples from \imagenet~\cite{krizhevsky2012imagenet} and \vtab{}~\cite{zhai2019vtab}.
    \item In Tables~\ref{tab:full_fgvc} and~\ref{tab:full_vtab}, we show per-task results for our SPT variants on FGVC and \vtab{} benchmarks, respectively.
    
\end{itemize}

\section{More Details of Contenders} 
\label{subsec:supp_contenders}

\begin{itemize}[leftmargin=2em]{

\item \fullft{}: fully tunes all the backbone and classification head parameters.
\vspace{-0.75em}
\item\linear{}: freezes all the backbone parameters and only tunes a linear classification head.
\vspace{-0.75em}
\item\bias{}~\cite{zaken2022bitfit}: freezes all the backbone parameters except for the bias terms and also tunes the linear classification head.
\vspace{-0.75em}
\item\partialft{}-$k$: freezes all the backbone parameters except for the last $k$ layers and also tunes the linear classification head as described in~\cite{jia2022vpt}.
\vspace{-0.75em}
\item \mlp{}-$k$: freezes all the backbone parameters and tunes the classification head which is implemented by a trainable $k$-layer multi-layer perceptron as described in~\cite{jia2022vpt}.
\vspace{-0.75em}
\item \shallowprompt{}~\cite{jia2022vpt}: freezes all the backbone parameters while introducing additional trainable prompts to the input space of the pretrained ViT.
\vspace{-0.75em}
\item \deepprompt{}~\cite{jia2022vpt}: freezes all the backbone parameters while appending additional trainable prompts to the sequence in the multi-head self-attention layer of each ViT block.
\vspace{-0.75em}
\item\adapter{}-$k$~\cite{houlsby2019parameter}: freezes all the backbone parameters while adding a down projection, a ReLU~\cite{hendrycks2016gaussian} non-linearity, and an up projection layer sequentially in the feed-forward network (FFN) of each visual Transformer block. 
We follow the training details of~\cite{zhang2022neural} to achieve better performance.
\vspace{-0.75em}
\item \lora{}-$k$~\cite{hu2022lora}: freezes all the backbone parameters while adding a concurrent branch including two low-rank matrices to the weight matrices in the multi-head self-attention layers to approximate efficiently updating them. 
The low-rank matrices can be merged into the backbone weights after fine-tuning. We follow the training details of~\cite{zhang2022neural} to achieve better performance.
\vspace{-0.75em}
\item \adaptformer{}~\cite{chen2022adaptformer}: freezes all the backbone parameters while adding a concurrent branch including a down projection, a ReLU~\cite{agarap2018deep} non-linearity, an up projection layer, and a pre-defined scaling factor to the FFN layer of each ViT block.
\vspace{-0.75em}
\item \noah{}~\cite{zhang2022neural}: searches for an optimal configuration with a once-for-all~\cite{cai2019once} network that includes trainable prompts, adapter modules, and LoRA modules, which requires a longer training schedule than the other VPET methods.
}
\end{itemize}

\section{More Parameter Sensitivity Patterns}
\label{subsec:supp_pattern1}
\begin{figure}[htb]
\begin{center}
    \includegraphics[width=\linewidth]{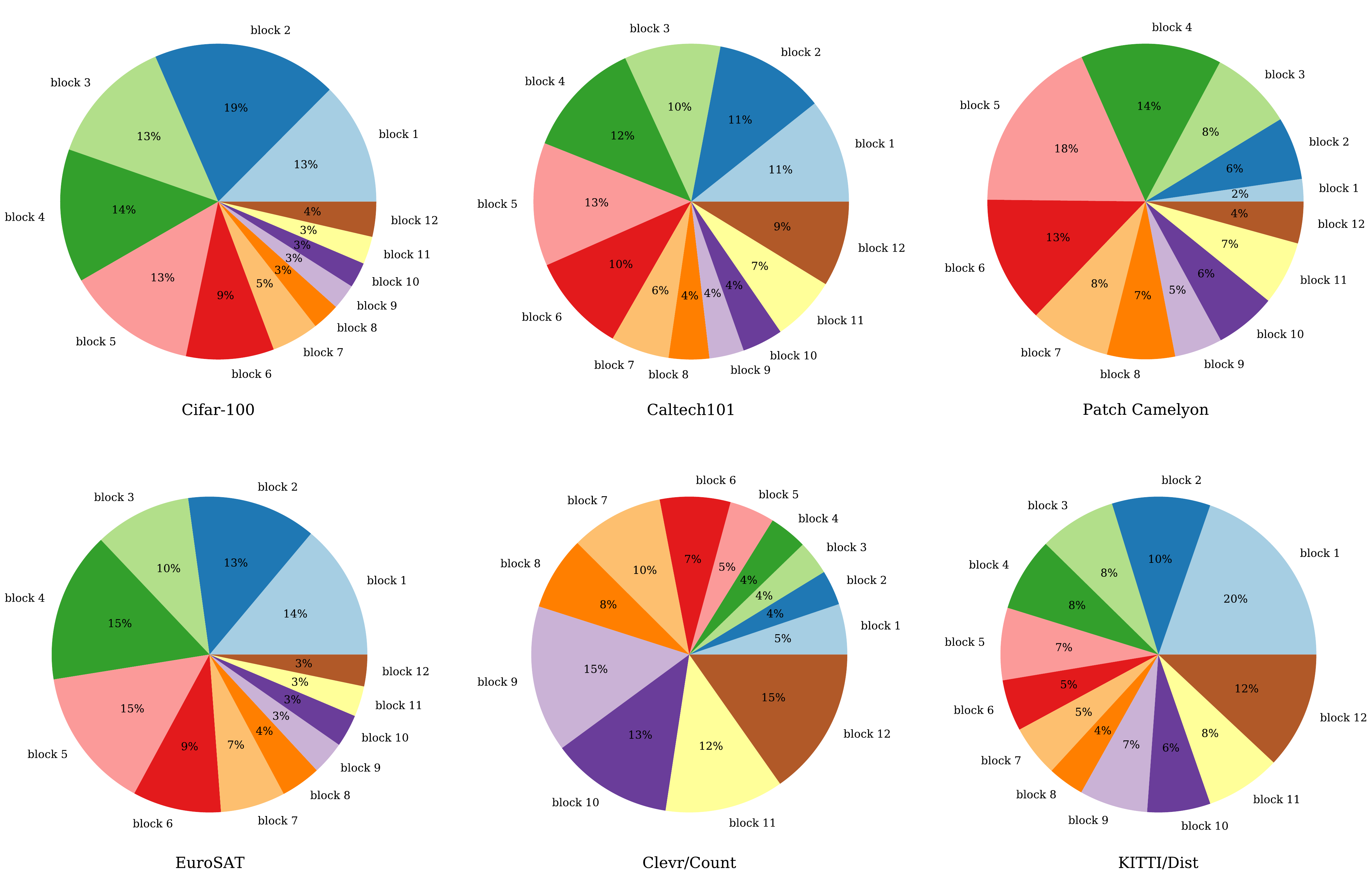}
\end{center}
\caption{The distribution of sensitive parameters by blocks under 0.4M trainable parameter budget with supervised pre-trained ViT-B/16 backbone. We sample six tasks from VTAB-1k~\cite{zhai2019vtab}.
}
\label{fig:sensitive_sup}
\end{figure}

\begin{figure}[htb]
\begin{center}
    \includegraphics[width=\linewidth]{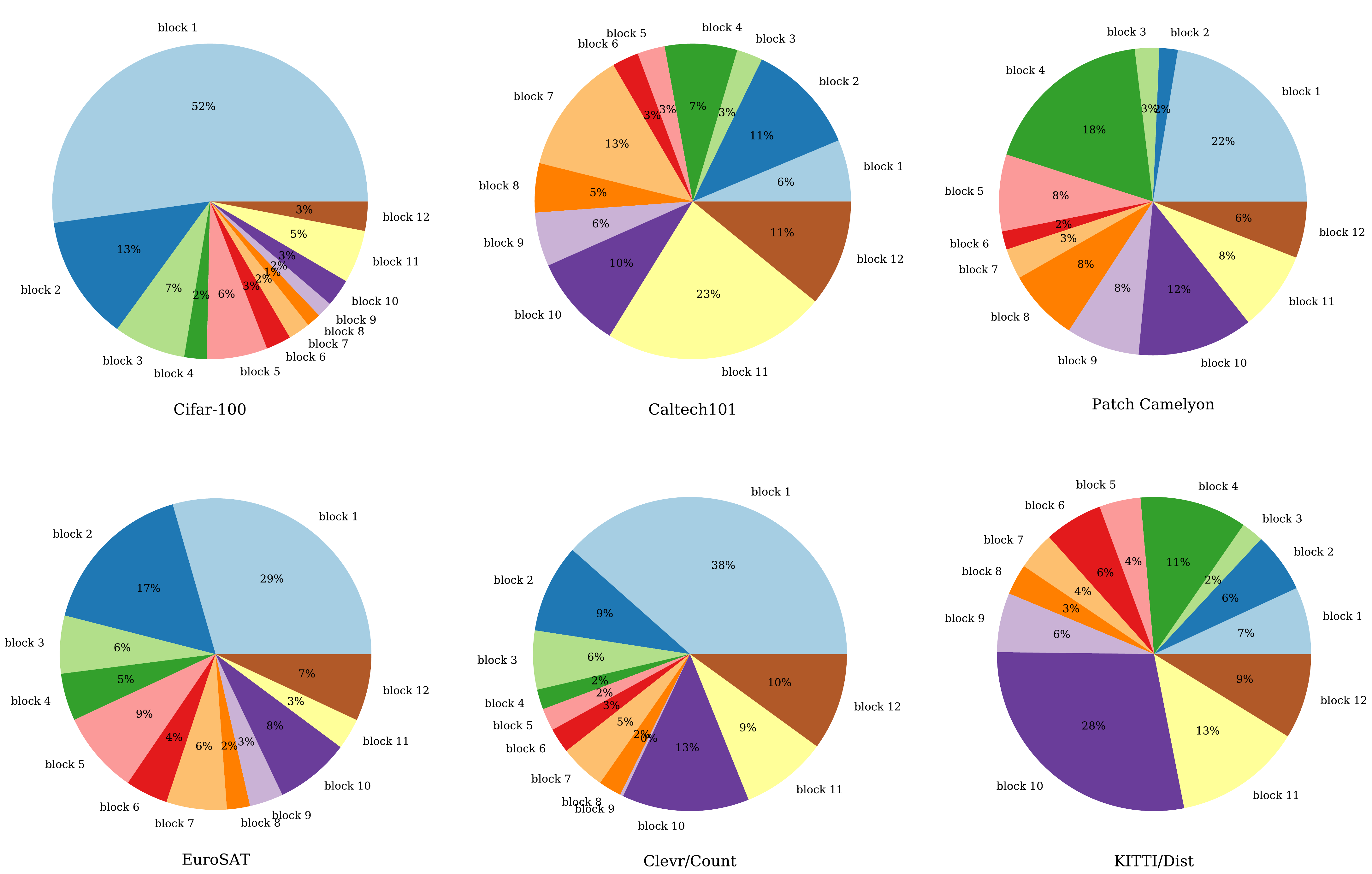}
\end{center}
\caption{The distribution of sensitive parameters by blocks under 0.4M trainable parameter budget with \mae{}~\cite{he2022masked} pre-trained ViT-B/16 backbone. We sample six tasks from VTAB-1k~\cite{zhai2019vtab}.}
\label{fig:sensitive_mae}
\end{figure}

\begin{figure}[tb]
\begin{center}
    \includegraphics[width=\linewidth]{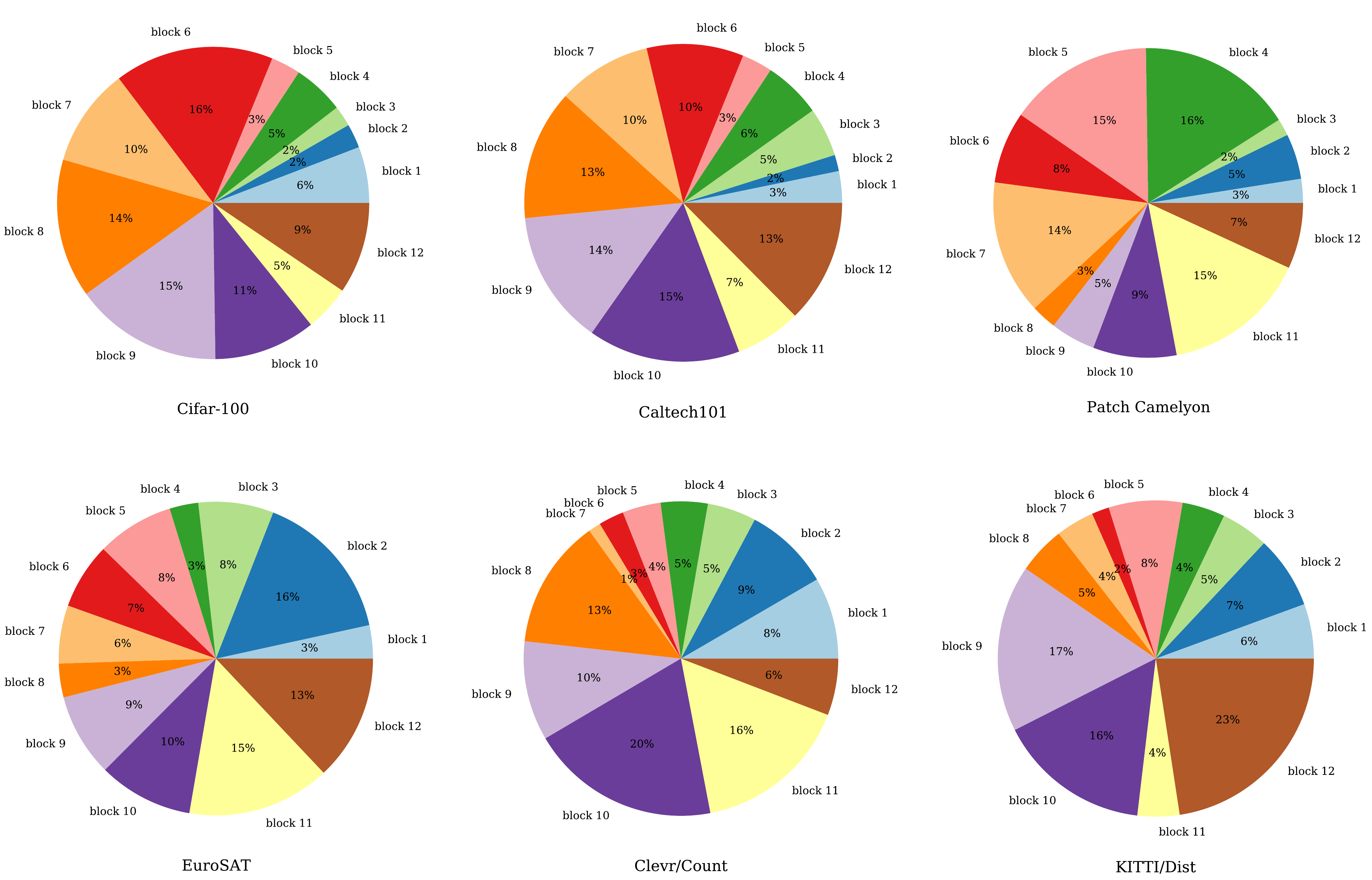}
\end{center}
\caption{The distribution of sensitive parameters by blocks under 0.4M trainable parameter budget for \moco{}~\cite{chen2021empirical} pre-trained ViT-B/16 backbone. We sample six tasks from VTAB-1k~\cite{zhai2019vtab}.}
\label{fig:sensitive_moco}
\end{figure}

\begin{figure}[tb]
\begin{center}
    \includegraphics[width=0.8\linewidth]{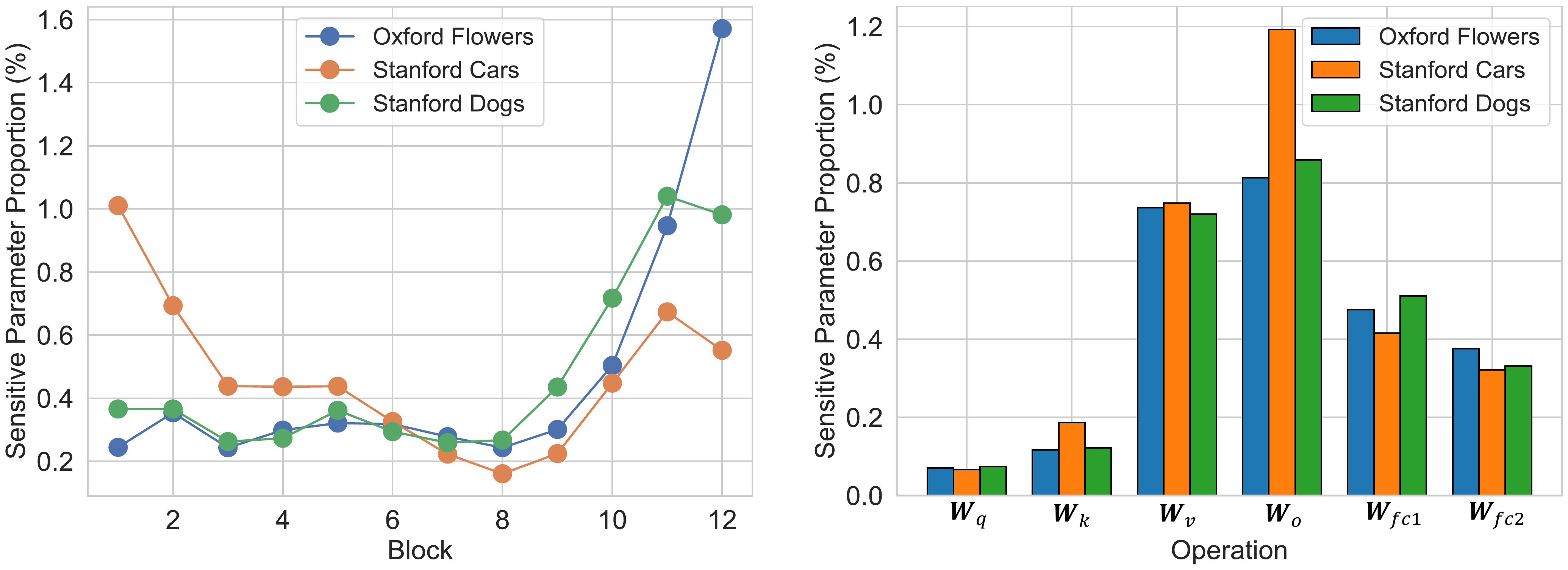}
\end{center}
\caption{Sensitivity patterns under 0.4M trainable parameters for Oxford Flowers~\cite{nilsback2008automated}, Stanford Cars~\cite{gebru2017cars}, and Stanford Dogs~\cite{Khosla_FGVC2011dogs}. We show the proportions of the sensitive
parameters for the query $\mW_{q}$, key $\mW_{k}$, value $\mW_{v}$, and $\mW_{o}$ weight matrices in the multi-head self-attention layer and two weight matrices $\mW_{fc1}$ and $\mW_{fc2}$ in the feed-forward network. 
}
\label{fig:sens_fgvc}
\end{figure}

\rev{We show more parameter sensitivity patterns for ViT-B/16 with various pre-training strategies (i.e., MAE~\cite{he2022masked} and MoCo V3~\cite{chen2021empirical}) and datasets sampled from FGVC benchmark~\cite{jia2022vpt}. We visualize the proportions of the sensitive parameters under 0.4M trainable parameter budget. Visualizations of sampled VTAB-1k datasets with MAE and MoCo V3 pre-trained ViT-B/16 are shown in Figures~\ref{fig:sensitive_sup},~\ref{fig:sensitive_mae},~\ref{fig:sensitive_moco}. Visualizations of sampled FGVC datasets with supervised pre-trained ViT-B/16 are shown in Figure~\ref{fig:sens_fgvc}. We find our observations in the main paper are general: the proportions of the sensitive parameter exhibit: 1) dataset-specific varying patterns in terms of network depth; and 2) dataset-agnostic similar patterns in terms of operations. We empirically find} that the self-supervised pre-trained backbones have higher sensitivity variances than the supervised pre-trained one across the 19 downstream tasks. In particular, the variance of ViT-B/16 pre-trained with MAE~\cite{he2022masked} is twice as large as that of the supervised pre-trained ViT-B/16. We speculate that our SPT variants can better handle the large variances for self-supervised pre-trained backbones (Table 2 of the main paper) by identifying task-specific positions to introduce the trainable parameters.

\begin{figure}[htb]
\begin{center}
    \includegraphics[width=0.6\linewidth]{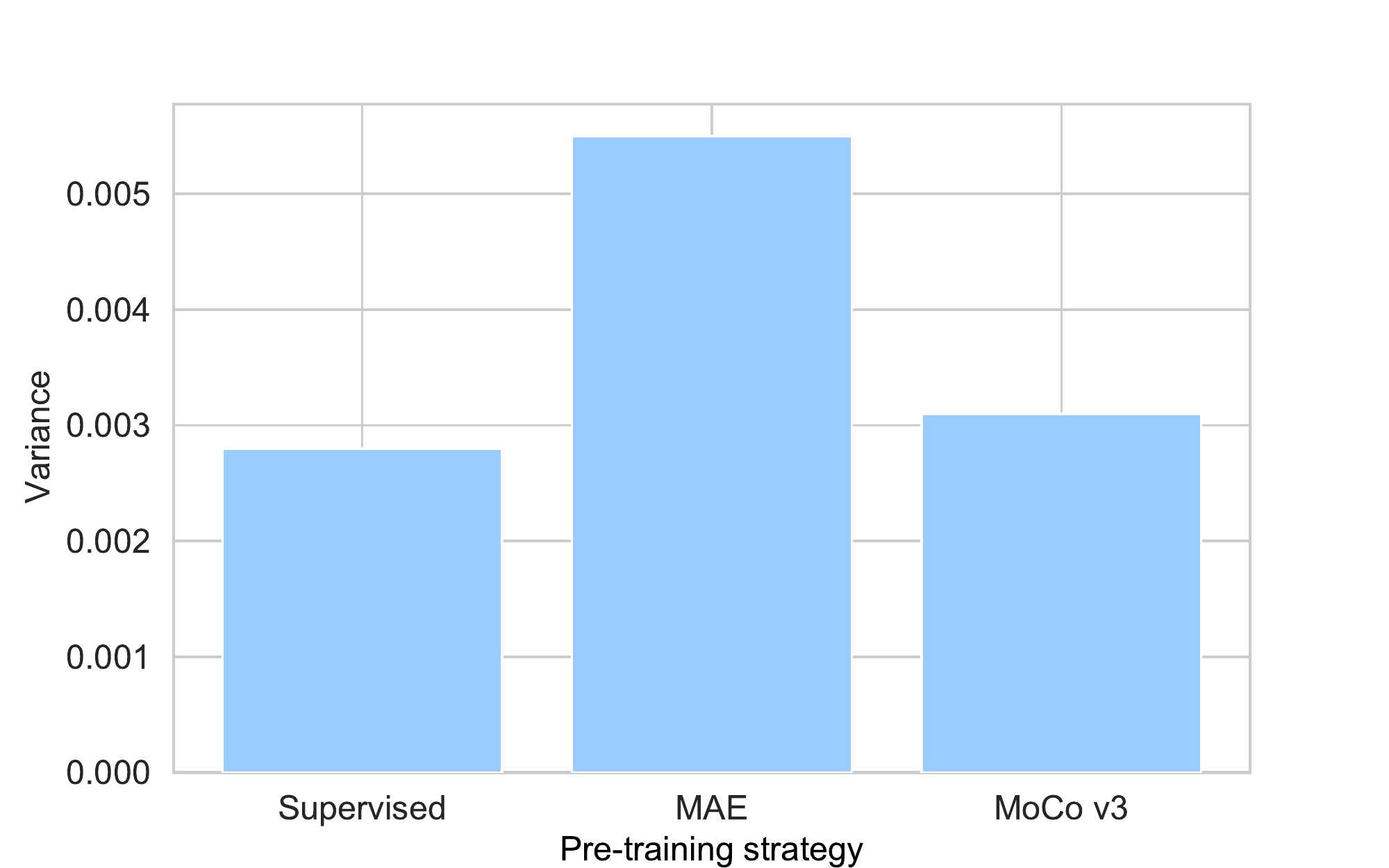}
\end{center}
\caption{Comparisons of sensitivity variances across backbones with different pre-training strategies on \vtab{}.}
\label{fig:variance}
\end{figure}

\begin{figure}[htb]
\begin{center}
\includegraphics[width=0.6\linewidth]{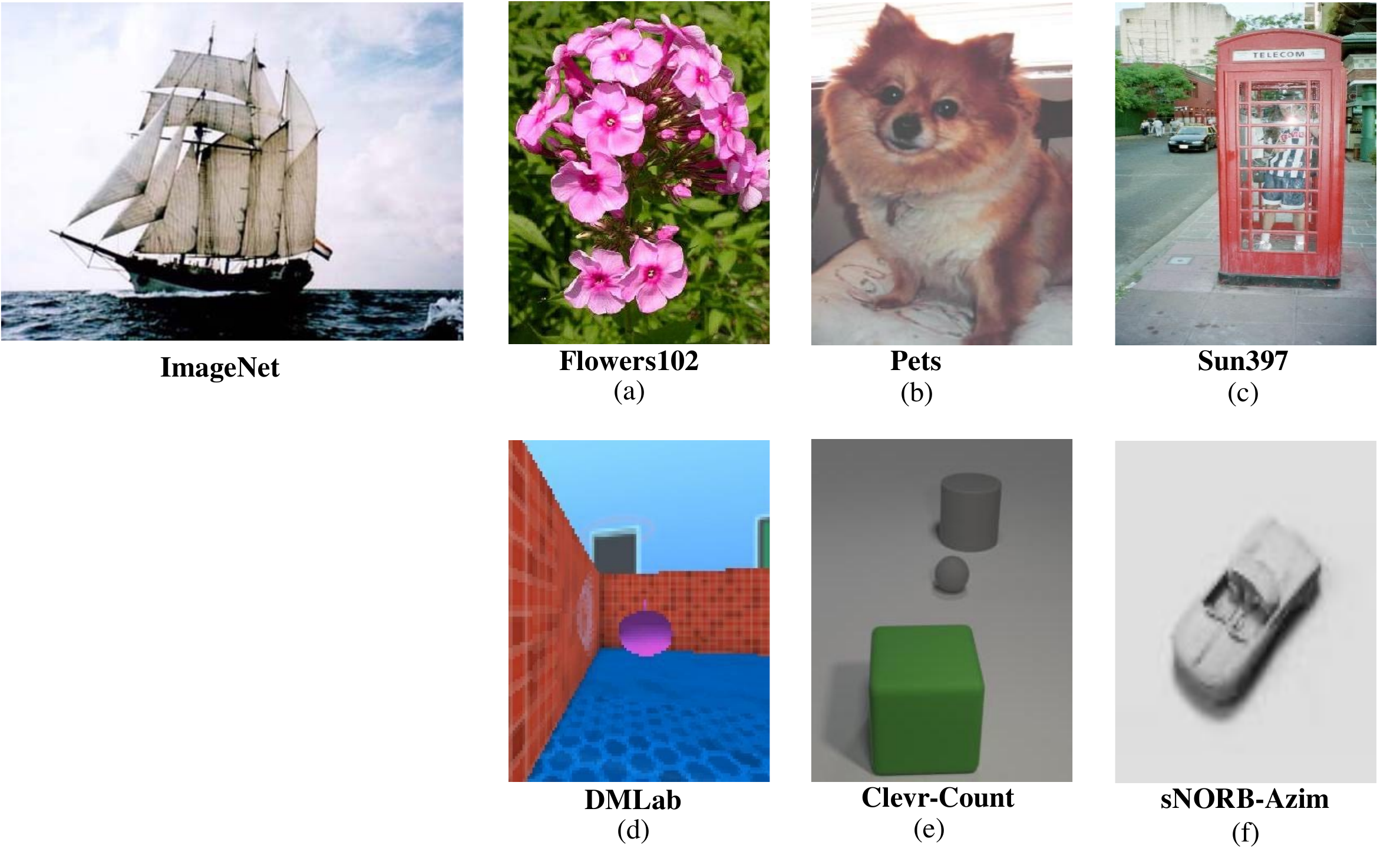}
\end{center}
\caption{Dataset samples from \imagenet~\cite{krizhevsky2012imagenet} and \vtab{}~\cite{zhai2019vtab}. Samples from Natural tasks of \vtab{} ((a), (b), and (c)) are relatively more similar to the source \imagenet{} samples compared to the ones from Structured tasks of \vtab{} ((d), (e), and (f)).}
\label{fig:domain}
\end{figure}

\section{Dataset Samples for the Source and Target Domains}
\label{subsec:supp_visual}
We visualize some sampled images from the source domain (\imagenet~\cite{krizhevsky2012imagenet}) and the target domains (\vtab{}~\cite{zhai2019vtab}) in Figure~\ref{fig:domain}. We observe that the images from the Natural tasks of \vtab{} are relatively more similar to the source domain compared to those from the Structured tasks of \vtab{}, which aligns with our observation that Structured tasks have large domain gaps. As structured tuning
improves the performance of Structured datasets (Section 4.3 of the main paper), we speculate that
structured tuning facilitates mitigating such large domain gaps.

\begin{table}[t]

\scriptsize
\resizebox{\textwidth}{!}{%
    \begin{tabular}{lc| cccccc}
    \toprule
      &  Tuned / Total &\bf{\cub{}} 
  &\bf{\nabirds{}}
  &\bf{\flowers{}} &\bf{\dogs{}} &\bf{\cars{}}
  &\bf{Mean Acc.} \\
    \midrule
    \band \fullft{} & 100\% &87.3 &82.7 &98.8 &89.4 &84.5 &88.5\\
    \midrule
        \multicolumn{8}{c}{\bf{Addition-based methods}}\\
    \midrule
    \mlp{}-3 & 1.50\% &85.1 &77.3 &97.9 &84.9 &53.8 &79.8
    \\
    \shallowprompt{} & 0.31\% & 86.7 &78.8 &98.4 &\underline{90.7} &68.7 &84.6\\
    \deepprompt{} & 0.98\% & \underline{88.5} &\underline{84.2} &\underline{99.0} &90.2 &83.6 &89.1\\
    \adapter{}-8 & 0.39\% & 87.3 &\textbf{84.3} &98.4 &88.8 &68.4 &85.5\\
    \adapter{}-32 & 0.95\% & 87.2 &\textbf{84.3} &98.5 &89.6 &68.4 &85.6\\
    \adaptformer{} & 0.44\% & 84.7 &75.2 &97.9 &84.7 &83.1 &85.1\\
    \SPTa{} & 0.41\% & \textbf{89.1} &83.3 &\textbf{99.2} &90.5 &\underline{85.6} &\underline{89.5}\\
    \SPTa{} & 0.47\% & \textbf{89.1} &83.3 &\textbf{99.2} &\textbf{91.1} &\textbf{86.2} &\textbf{89.8}\\
    \midrule
     \multicolumn{8}{c}{\bf{Reparameterization-based methods}}\\
    \midrule
    \linear{} & 0.12\% & 85.3 &75.9 &97.9 &86.2 &51.3 &79.3\\
    \partialft{}-1 & 8.38\% &85.6 &77.8 &98.2 &85.5 &66.2 &82.6\\
    \bias{} & 0.13\% &\underline{88.4} &\textbf{84.2} &98.8 &\underline{91.2} &79.4 &88.4\\
    \lora{}-8 & 0.55\% &84.9 &79.0 &98.1 &88.1 &79.8 &86.0 \\
    \lora{}-16 & 0.90\% &85.6 &79.8 &98.9 &87.6 &72.0 &84.8 \\
    \SPTl{} & 0.41\% &\textbf{88.6} &82.8 &\underline{99.4} &\textbf{91.4} &\underline{84.5} &\underline{89.3} \\
    \SPTl{} & 0.60\% &\textbf{88.6} &\underline{83.4} &\textbf{99.5} &\textbf{91.4} &\textbf{87.3} &\textbf{90.1} \\
\bottomrule
    \end{tabular}}
    \caption{
    Per-task results on the FGVC benchmark from Table~1 of the main paper. ``Tuned / Total'' denotes the fraction of the trainable parameters. Top-1 accuracy (\%) is reported. The best result is in \textbf{bold}, and the second-best result is \underline{underlined}.
}\label{tab:full_fgvc}
\end{table}

\begin{sidewaystable}[t]
\scriptsize
\resizebox{\textwidth}{!}{%
    \begin{tabular}{lc | cccccccc | ccccc | ccccccccc}
    \toprule
    & & \multicolumn{8}{c|}{\textbf{Natural}} & \multicolumn{5}{c|}{\textbf{Specialized}} & \multicolumn{9}{c}{\textbf{Structured}} \\
    & \rotatebox{90}{Tuned / Total} & \rotatebox{90}{\bf{Cifar100}} & \rotatebox{90}{\bf{Caltech101}} & \rotatebox{90}{\bf{DTD}} & \rotatebox{90}{\bf{Flower102}} & \rotatebox{90}{\bf{Pets}} & \rotatebox{90}{\bf{SVHN}}  & \rotatebox{90}{\bf{Sun397}} & \rotatebox{90}{\bf{Mean Acc.}} & \rotatebox{90}{\bf{Camelyon}}  & \rotatebox{90}{\bf{EuroSAT}}   & \rotatebox{90}{\bf{Resisc45}}  & \rotatebox{90}{\bf{Retinopathy}} & \rotatebox{90}{\bf{Mean Acc.}} & \rotatebox{90}{\bf{Clevr-Count}} & \rotatebox{90}{\bf{Clevr-Dist}}  & \rotatebox{90}{\bf{DMLab}} & \rotatebox{90}{\bf{KITTI-Dist}}  & \rotatebox{90}{\bf{dSpr-Loc}} & \rotatebox{90}{\bf{dSpr-Ori}}   & \rotatebox{90}{\bf{sNORB-Azim}}  & \rotatebox{90}{\bf{sNORB-Ele}} & \rotatebox{90}{\bf{Mean Acc.}}   \\
    \midrule
\band \fullft{} & 100\% &68.9 &87.7 &64.3 &97.2 &86.9 &87.4 &38.8 &75.9 &79.7 &95.7 &84.2 &73.9 &83.4 &56.3 &58.6 &41.7 &65.5 &57.5 &46.7 &25.7 &29.1 &47.6

    \\\midrule
     \multicolumn{22}{c}{\bf{Addition-based methods}}
    \\\midrule
    \mlp{}-3 & 1.50\% &63.8 &84.7 &62.3 &97.4 &84.7 &32.5 &49.2 &67.8 &77.0 &88.0 &70.2 &56.1 &72.8 &47.8 &32.8 &32.3 &58.1 &12.9 &21.2 &15.2 &24.8 &30.6\\
    \shallowprompt{} & 0.31\%  & 77.7 &86.9 &62.6 &97.5 &87.3 &74.5 &51.2 &76.8 &78.2 &92.0 &75.6 &72.9 &79.7 &50.5 &58.6 &40.5 &67.1 &68.7 &36.1 &20.2 &34.1 &47.0\\
    \deepprompt{} & 0.98\% &78.8 &90.8 &65.8 &98.0 &88.3 &78.1 &49.6 &78.5 &81.8 &96.1 &83.4 &68.4 &82.4 &68.5 &60.0 &46.5 &72.8 &73.6 &47.9 &32.9 &37.8 &55.0\\
    \adapter{}-8  & 0.39\% & 69.2 & 90.1 & 68.0 & 98.8 & 89.9 & 82.8 & 54.3 & 79.0 & 84.0 & 94.9 & 81.9 & 75.5 & 84.1 & 80.9 & 65.3 & 48.6 & 78.3 & 74.8 & 48.5 & 29.9 & 41.6 & 58.5\\
    \adapter{}-32 & 0.71\% & 68.7 & 92.2 & 69.8 &98.9 & 90.3& 84.2& 53.0& 79.6& 83.2& 95.4& 83.2& 74.3 & 84.0 & 81.9 & 63.9& 48.7 & 80.6& 76.2& 47.6& 30.8& 36.4 & 58.3 \\
    \noah{} & 0.50\% & 69.6 & 92.7 & 70.2 & 99.1 & 90.4 & 86.1 & 53.7 & 80.2 & 84.4 & 95.4 & 83.9 & 75.8 & 84.9 & 82.8 & 68.9 & 49.9 & 81.7 & 81.8 & 48.3 & 32.8 & 44.2 & 61.3\\
    \SPTa{} & 0.30\% & 72.9 & 93.2 & 72.5 & 99.3 & 91.4 & 84.6 & 55.2 & 81.3 & 85.3 & 96.0 & 84.3 & 75.5 & 85.3 & 82.2 & 68.0 & 49.3 & 80.0 & 82.4 & 51.9 & 31.7 & 41.2 & 60.8\\
    \SPTa{} & 0.44\% & 72.9 & 93.2 & 72.5 & 99.3 & 91.4 & 88.8 & 55.8 & 82.0 & 86.2 & 96.1 & 85.5 & 75.5 & 85.8 & 83.0 & 68.0 & 51.9 & 81.2 & 82.4 & 51.9 & 31.7 & 41.2 & 61.4\\
    \midrule
     \multicolumn{22}{c}{\bf{Reparameterization-based methods}}
    \\\midrule
    \linear{} & 0.12\% & 63.4 &85.0 &63.2 &97.0 &86.3 &36.6 &51.0 &68.9 &78.5 &87.5 &68.6 &74.0 &77.2 &34.3 &30.6 &33.2 &55.4 &12.5 &20.0 &9.6 &19.2 &26.8\\
    \partialft{}-1 & 8.38\% &66.8 &85.9 &62.5 &97.3 &85.5 &37.6 &50.6 &69.4 &78.6 &89.8 &72.5 &73.3 &78.5 &41.5 &34.3 &33.9 &61.0 &31.3 &32.8 &16.3 &22.4 &34.2\\
    \bias{} & 0.13\% &72.8 &87.0 &59.2 &97.5 &85.3 &59.9 &51.4 &73.3 &78.7 &91.6 &72.9 &69.8 &78.3 &61.5 &55.6 &32.4 &55.9 &66.6 &40.0 &15.7 &25.1 &44.1\\
\lora{}-8 & 0.55\% & 67.1 & 91.4 & 69.4 & 98.8 & 90.4 & 85.3 & 54.0 &79.5 & 84.9 & 95.3 & 84.4 & 73.6 & 84.6 & 82.9 & 69.2 & 49.8 & 78.5 & 75.7 & 47.1 & 31.0 & 44.0 & 60.5 \\
\lora{}-16 & 0.90\% & 68.1 & 91.4 & 69.8 & 99.0 & 90.5 & 86.4 & 53.1 &79.8 & 85.1 & 95.8 & 84.7 & 74.2 & 84.9 & 83.0 & 66.9 & 50.4 & 81.4 & 80.2 & 46.6 & 32.2 & 41.1 & 60.2 \\
\SPTl{} & 0.31\% & 72.3 & 93.0 & 72.5 & 99.3 & 91.5 & 86.2  & 55.5 & 81.5 & 85.0 & 96.2 & 85.1 & 75.9 & 85.6 & 83.7 & 66.4 & 52.5 & 80.2 & 80.1 & 51.1 &  30.1 & 41.3 & 60.7 \\
\SPTl{} & 0.63\% & 73.5 & 93.3 & 72.5 & 99.3 & 91.5 & 87.9 & 55.5 & 81.9 & 85.7 & 96.2 & 85.9 & 75.9 & 85.9 & 84.4 & 67.6 & 52.5 & 82.0 & 81.0 & 51.1 &  30.2 & 41.3 & 61.3 \\
\bottomrule
    \end{tabular}}
    \caption{
    Per-task results on the \vtab{} benchmark from Table~1 of the main paper. ``Tuned / Total'' denotes the fraction of the trainable parameters. Top-1 accuracy (\%) is reported.
}\label{tab:full_vtab}
\end{sidewaystable}

\end{document}